%% file: main.tex
\documentclass[letterpaper]{article} % DO NOT CHANGE THIS
\usepackage{aaai24}  % DO NOT CHANGE THIS
\usepackage{times}  % DO NOT CHANGE THIS
\usepackage{helvet}  % DO NOT CHANGE THIS
\usepackage{courier}  % DO NOT CHANGE THIS
\usepackage[hyphens]{url}  % DO NOT CHANGE THIS
\usepackage{graphicx} % DO NOT CHANGE THIS
\usepackage{natbib}  % DO NOT CHANGE THIS AND DO NOT ADD ANY OPTIONS TO IT
\usepackage{caption} % DO NOT CHANGE THIS AND DO NOT ADD ANY OPTIONS TO IT
\usepackage{algorithm}
\usepackage{algorithmic}

\input{custom_shortcut}

\usepackage{newfloat}
\usepackage{listings}
\DeclareCaptionStyle{ruled}{labelfont=normalfont,labelsep=colon,strut=off} % DO NOT CHANGE THIS
\floatstyle{ruled}
\newfloat{listing}{tb}{lst}{}
\floatname{listing}{Listing}
\usepackage{epsfig}
\usepackage{booktabs} 
\usepackage{amsfonts}
\usepackage{amsmath}
\usepackage{amssymb}
\usepackage{mathrsfs} 
\usepackage{array,multirow}
\usepackage[dvipsnames]{xcolor}
\usepackage{dirtytalk}
\usepackage{hhline}
\usepackage{colortbl}
\usepackage{float}
\definecolor{limegreen}{HTML}{DBEAC1}
\definecolor{lavender}{HTML}{EEE2F8}
\definecolor{lightorange}{HTML}{FFF7F4}
\definecolor{lightgray}{HTML}{767070}
\definecolor{lightblue}{HTML}{DAE3F4}
\definecolor{beige}{HTML}{F3E5AB}
\usepackage{nicefrac}       % compact symbols for 1/2, etc.
\usepackage{microtype}      % microtypography

\title{Hyp-OW: Exploiting Hierarchical Structure Learning with Hyperbolic Distance Enhances Open World Object Detection}
\author{
    Thang Doan\thanks{corresponding author} \hspace{5mm}  Xin Li \hspace{5mm}  Sima Behpour \hspace{5mm} 
    Wenbin He \hspace{5mm}   Liang Gou \hspace{5mm}   Liu Ren 
}
\affiliations{
    Bosch Research North America $\&$  Bosch Center for Artificial Intelligence (BCAI)\\
     \texttt{\{thang.doan,xin.li9,sima.behpour,wenbin.he2,liang.gou,liu.ren\}@us.bosch.com}
}

\usepackage{bibentry}
\begin{document}

\maketitle

\begin{abstract}
Open World Object Detection (OWOD) is a challenging and realistic task that extends beyond the scope of standard Object Detection task. It involves detecting both known and unknown objects while integrating learned knowledge for future tasks. However, the level of "unknownness" varies significantly depending on the context. For example, a tree is typically considered part of the background in a self-driving scene, but it may be significant in a household context. We argue that this contextual information should already be embedded within the known classes. In other words, there should be a semantic or latent structure relationship between the known and unknown items to be discovered. Motivated by this observation, we propose Hyp-OW, a method that learns and models hierarchical representation of known items through a SuperClass Regularizer. Leveraging this representation allows us to effectively detect unknown objects using a similarity distance-based relabeling module. Extensive experiments on benchmark datasets demonstrate the effectiveness of Hyp-OW, achieving improvement in both known and unknown detection (up to 6 percent). These findings are particularly pronounced in our newly designed benchmark, where a strong hierarchical structure exists between known and unknown objects. Our code can be found at \textcolor{orange}{https://github.com/boschresearch/Hyp-OW}.
\end{abstract}

\input{1-introduction.tex}

\input{2-related_works.tex}

\input{3-background.tex}

\input{4-methodology.tex}

\input{5-experiments.tex}
\input{6-conclusion.tex}

\clearpage
\bibliography{aaai24}

\clearpage
\appendix
\input{appendix.tex}

\end{document}

%% file: custom_shortcut.tex
\newcommand{\cA}{\mathcal{A}}
\newcommand{\cB}{\mathcal{B}}

\newcommand{\cD}{\mathcal{D}}

\newcommand{\cI}{\mathcal{I}}

\newcommand{\cK}{\mathcal{K}}
\newcommand{\cL}{\mathcal{L}}
\newcommand{\cM}{\mathcal{M}}

\newcommand{\cP}{\mathcal{P}}

\newcommand{\cS}{\mathcal{S}}
\newcommand{\cT}{\mathcal{T}}
\newcommand{\cU}{\mathcal{U}}

\newcommand{\cY}{\mathcal{Y}}

\newcommand{\mb}{\mathbf}

\newcommand{\method}{Hyp-OW }

\newcommand{\lowregime}{\textcolor{Periwinkle}{Low regime } }
\newcommand{\mediumregime}{\textcolor{LimeGreen}{Medium regime }}
\newcommand{\highregime}{\textcolor{Red}{High regime }}

%% file: 1-introduction.tex
\section{Introduction}

Advances in Object Detection (OD) have unlocked a plethora of practical applications such as robotics ~\cite{object_detection_robotics}, self-driving cars ~\cite{object_detection_self_driving_cars}, manufacturing ~\cite{object_detection_manufacturing}, and medical analysis ~\cite{object_detection_medical_analysis}. Recent breakthroughs in attention-based neural network architecture, such as Deformable Transformers ~\cite{deformable_transformer}, have yielded impressive performance in these settings. However, most of these approaches assume a fixed number of classes (closed-world assumption), which is rare in reality. Continual Object Detection ~\cite{continual_od} takes a step further by incrementally adding new classes, resulting in a distribution shift in the input and the well-known phenomenon of \textit{catastrophic forgetting} \citep{Kirkpatrick,pmlr-v130-ntk_catastrophic} where the network forgets previously learned knowledge. Open World (OW) \cite{open_world_recognition_2015} takes these assumptions even further, introducing the detection and integration of newly discovered classes.

While the seminal work by ~\cite{open_world_recognition_2015} introduced OW framework, further advancements by ~\cite{owod} extended it in two key aspects: the detection task and continual learning. However, a significant challenge within this framework lies in the absence of annotations for unknown objects, leading to biases toward known labels and potential confusion between unknown items and the background. This bias significantly impedes the accurate identification of unknown objects and presents a major hurdle in the detection process.

Previous approaches, often relying on shared features or objectness scores~\cite{owod,owdetr,prob}, as well as clustering methods~\citep{uc_owod,owod_discriminative_prototypes}, have failed to address a critical challenge: defining what constitutes an "unknown" object. Currently, there is no clear definition or prior knowledge available to effectively distinguish unknowns from the background. Its interpretation greatly varies depending on the context. For example, in a driving scene, a "debris on the road" could be considered an unknown object ~\cite{object_detection_self_driving_cars}, while in a camera surveillance context, it might be perceived as part of the background ~\cite{obj_detection_camera_surveillance}. Without considering the context, these works can only learn to differentiate knowns and unknowns at low level features such as texture or shape. As a consequence, they fail to model any hierarchical structures and similarities between known and unknown items, whether at the image level or dataset level.

Acknowledging this context information, we argue that a hierarchical structure must exist between the objects to be discovered and the known items ~\cite{more_practical_owod}. This hierarchy is characterized by classes that share the same semantic context, belonging to the same category such as vehicles, animals, or electronics. Such hierarchical relationships enable the retrieval of common features and facilitate the discovery of unknown objects. For instance, a model trained on objects related to driving scenes can adequately detect stop signs or traffic lights but is not expected to recognize unrelated objects like a couch or any furniture.

Given this discrepancy, we propose modeling hierarchical relationships among items to enhance the discovery of unknowns. Ideally, items belonging to the same family (or category) should be closer to each other while being further away from different families (e.g., animals versus vehicles). To capture these structures, Hyperbolic Distance~\citep{nickel2018learning,park2021unsupervised}, which naturally maps hierarchical latent structures, such as graphs or trees, emerges as an ideal distance metric. This has the desirable property of capturing the affinity between unknown items and known items, thereby enhancing the detection of unknown objects.

\begin{figure}[h!]
 
    \includegraphics[width=1.0\linewidth]{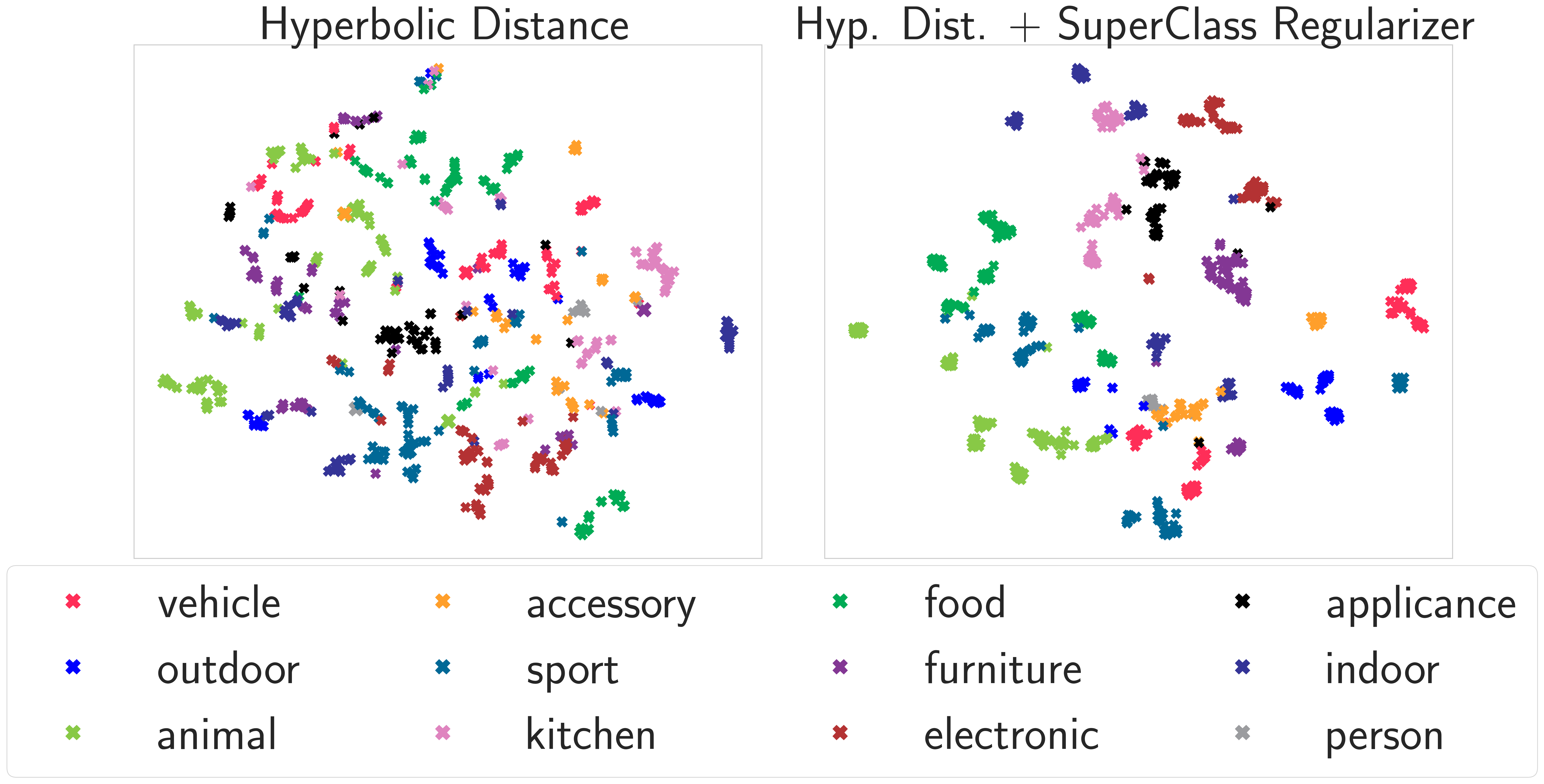}
  
  \caption{\textbf{t-SNE plot of the learned class representations, with colors representing their respective categories.} Our SuperClass Regularizer (right) learns the hierarchical structure by grouping together classes from the same category while pushing apart those from different categories.}
  \label{fig:tse_mini}
\end{figure}

\paragraph{Contribution}
Motivated by the aforementioned literature gap, we propose a \textbf{\underline{Hyp}}erbolic Distance-based Adaptive Relabeling Scheme for \textbf{\underline{O}}pen \textbf{\underline{W}}orld Object Detection (Hyp-OW). Our contribution can be summarized in three parts: 
\begin{itemize}
\item  \method is a simple yet effective method that learns innate hierarchical structure between objects grouping item from the same category closer while pushing classes from different categories further apart through a SuperClass Regularizer (illustrated in Figure~\ref{fig:tse_mini}, right).
\item We propose an Adaptive Relabeling Scheme that enhances the detection of unknown objects by leveraging the semantic similarity between known and unknown objects in the hyperbolic space. 
 
\item Our experiments demonstrate significant improvements in both unknown recall (up to $6\%$) and known object detection performance (up to $5\%$) with \method. These gains are particularly prominent when evaluating on our (designed) Hierarchical dataset that exhibits high inherent hierarchical structures.

\end{itemize}

%% file: 2-related_works.tex
\section{Related Work}

\subsection{Open World Object Detection}
The OWOD framework, introduced by \cite{owod}, has inspired many recent works due to its realistic and close-to-real-world setting that integrates newly discovered items into the base knowledge progressively. While the first stream of work was originally based on the Faster-RCNN model \citep{owod,owod_discriminative_prototypes,uc_owod,two_branch}, more recent works have utilized Deformable Transformers due to their superior performance \citep{owdetr,prob}. \citet{owod} introduced ORE, a Faster-RCNN-based model that learns class prototypes using contrastive learning with Euclidean distance. However, their approach relied on a held-out validation set where unknown items are explicitly labeled to learn an energy-based model to discriminate unknown items. \cite{owod_discriminative_prototypes} extended this setting by minimizing the overlap between the distributions of unknown and known classes. OW-DETR \citep{owdetr} designed a novelty-branch head to relabel the top-k highest background scores as unknowns. These pseudo-labels relied on unmatched bounding box proposals with high backbone activation being selected as unknown objects. On the other hand, \citet{two_branch} decoupled the localization and classification tasks (introduced by \citet{kim2021oln}) by learning a class-free head to localize objects. Recently, PROB \citep{prob} learned a probabilistic objectness score by learning common statistics for all objects using Mahalanobis distance \citep{lee2018simple} and considered all the remaining bounding box proposals as unknown items. During the evaluation phase, they filter out proposal bounding boxes using the latter probabilistic models.
% \thang{they learn an average objectness by averaging all the known class but leverage similarity of different known classes}

\subsection{Class-Agnostic Object Detection}

Another stream of work in the field of object detection is dubbed class-agnostic object detection, which focuses on localizing objects ~\citep{kim2021oln, two_branch, jaiswal2021class}. The objective is to remove the class label information and learn a shared low-level feature representation that effectively captures the essence of an object. ~\citet{kim2021oln} designed a pure localization head by introducing a second branch that is decoupled from the classification head. ~\citet{jaiswal2021class} introduced an adversarial objective loss function that penalizes label information in the encoded features. Pixel-wise class-free object detection ~\cite{pixel_wise_class_free_od} used texture gray level quantization to retrieve objects. Saito et al. \citet{saito2022learning} designed a new data augmentation method that pastes an annotated object onto an object-free background.  ~\citet{maaz2022class} leveraged language models to improve unknown detection with their Multi-Modal Vision Transformers.

\subsection{Learning Hierarchical Representation with Hyperbolic Distance}
Poincare embeddings have been widely used in the literature to learn hierarchical structures from complex symbolic or multi-relational data, which can be represented by graphs or trees, such as social networks or taxonomies ~\citep{nickel2018learning,pmlr-v97-law19a}. Due to its good performance, it has been applied to image classification as well ~\citep{Khrulkov_2020_CVPR,yan2021unsupervised,yue2023hyperbolic,ermolov2022hyperbolic}. For example, ~\citet{yan2021unsupervised} used hierarchical clustering to approximate a multi-layered tree structure representation that guides the hyperbolic distance learning process. Similarly, \citet{liu2020hyperbolic} used taxonomy embedding from GloVe ~\citep{pennington-etal-2014-glove} to learn a finer-grained representation. Hyperbolic distance has also been used for object detection ~\citep{lang2022hyperbolic,ge2022hyperbolic}. ~\citet{ge2022hyperbolic} was interested in learning context-object association rules by reasoning on different image scales. However, none of them leveraged the learned hyperbolic distance to retrieve unknowns items for OWOD.

%% file: 3-background.tex
\section{Background}

\subsection{Problem Formulation}
OWOD framework describes the setting where a user receives over the time a stream of $T$ tasks indexed by $t \in [1,T]$. Every task $t$ contains $C_t \in \mathbb{N}^{*}$ known classes (denoted by set $\cK^{t}$\footnote{Whenever there is no ambiguity we will remove the task index $t$ to de-clutter the notation.}). The goal is to train an object detector module $f$ to accurately recognize the known classes but also discovering unknown classes (denoted by set $\cU^{t}$). At the end of task $t$, $C_{t+1}$ unknown classes are labelled (with an oracle) and included in the next task $t+1$. The process repeat until task $T$ that does not contain anymore unknowns.

The dataset of task $t$ is defined as $\cD^{t}=\{ \cI^{t},\cY^{t}\}$ where $\cI^{t}$ are image inputs and $\cY^{t}$ the corresponding labels. Each label consists of a list of bounding box locations along with their corresponding labels. We follow the setting of OWOD \cite{owod} where a set of $K$ examplars of each class is stored in a replay buffer at the end of each task $t$ (to mitigate forgetting) to be replayed. Additionally, throughout the training, we will be storing item in a replay buffer $\cM$ with a capacity of $m$ exemplar per class. We denote $\cB$ the incoming batch.

\subsection{Deformable Transformers for OWOD}
We adopt Deformable Transformers \citep{deformable_transformer} as our base detector, as it showed simplicity and high performance~\cite{owdetr}. The image input is processed through a set of encoder-decoder modules to output $Q$ queries $\{\mb{q_{i}}\}_{i=1}^{Q}$, where $\mb{q_{i}} \in \mathbb{R}^{d}$ are the output embeddings. These queries served as input to different heads such as classification and localization heads. since $Q$ is higher than the number of ground-truth labels, the Hungarian algorithm~\citep{Kuhn1955Hungarian} is used to match the labeled ground-truth items with each query. We refer the reader to~\citep{deformable_transformer} for more details.

\subsection{Hyperbolic Embeddings}
A Hyperbolic space is a $n$-dimensional Riemann manifold defined as $(\mathbb{B}_{c}^{n},g^{\mathbb{M}})$
with its Poincare ball $\mathbb{B}_{c}^{n}=\{ x \in \mathbb{R}^{n}: c \lVert x \rVert^{2} \leq 1 , c \geq 0   \}$ ($c$ being the constant curvature) and equipped with a Riemannian metric $g^{\mathbb{M}}=(\frac{2}{1- \lVert \mb{x} \rVert^{2} })^{2}g^{E}$ where $g^{E}=\mathbf{I}_{n}$ is the Euclidian metric tensor. The transformation from the Euclidian to hyperbolic space is done via a bijection termed \textit{exponential} mapping $\exp_{\mb{b}}^{c}: \mathbb{R}^{n}  \rightarrow \mathbb{B}_{c}^{n}$. 
\begin{align}
\exp_{\mb{b}}^{c}(\mb{x})=\mb{b} \oplus_{c} (\tanh{(\sqrt{c}\frac{\lambda_{\mb{b}}^{c} \lVert \mb{x}  \rVert}{2})}\frac{\lVert \mb{x}  \rVert}{\sqrt{c}\lVert \mb{x}  \rVert})
\end{align}
with $\lambda_{\mb{b}}^{c}=\frac{2}{1-c \lVert \mb{b} \rVert^{2} }$ and the base point $\mb{b}$. The latter is often empirically taken as $\mb{b}=\mb{0}$ to simplify the formulas without impacting much the results \citep{ermolov2022hyperbolic}. We will also adopt this value in our study.

Inside this hyperbolic space, the distance between two points $\mb{x},\mb{y}  \in \mathbb{B}_{c}^{n}$ is computed as:
\begin{align}
d_{hyp}(\mb{x},\mb{y})=\frac{2}{\sqrt{c}}\arctan{(\sqrt{c}\lVert -\mb{x} \oplus_{c} \mb{y}  \rVert)}
\end{align}
where the addition operation $\oplus_{c} $ is defined as : \\
$\mb{x} \oplus_{c} \mb{y}=\frac{(1+2c \left< \mb{x}, \mb{y} \right> + c \lVert \mb{y}  \rVert^{2} )\mb{x} + (1 - c \lVert \mb{x}  \rVert^{2} )\mb{y} }{1+2c \left< \mb{x}, \mb{y} \right> + c^{2} \lVert \mb{x}  \rVert^{2} \lVert \mb{y}  \rVert^{2} }$.

From now on, we will denote $\mb{z_i}$ the projection of the queries $\mb{q_i}$ into the hyperbolic embedding space, i.e, $\mb{z_{i}}=\exp_{\mb{b}}^{c}(\mb{q_{i}})$. 
When $c \to 0$, we recover the Euclidian distance: $\lim_{c \to 0} d_{hyp}(\mb{x},\mb{y}) = 2 \rVert \mb{x} - \mb{y} \rVert$. This quantity is also related to the cosine similarity $d_{cos}(\mb{x},\mb{y})=2-2\frac{ < \mb{x},\mb{y} >}{\lVert \mb{x}  \rVert \cdot \lVert \mb{y}  \rVert}$ in the case of normalized vectors (See supplementary).

%% file: 4-methodology.tex
\section{\method}
\begin{figure*}[ht!]
  % \hspace{-10mm}
    \begin{center}
    \includegraphics[width=0.8\linewidth]{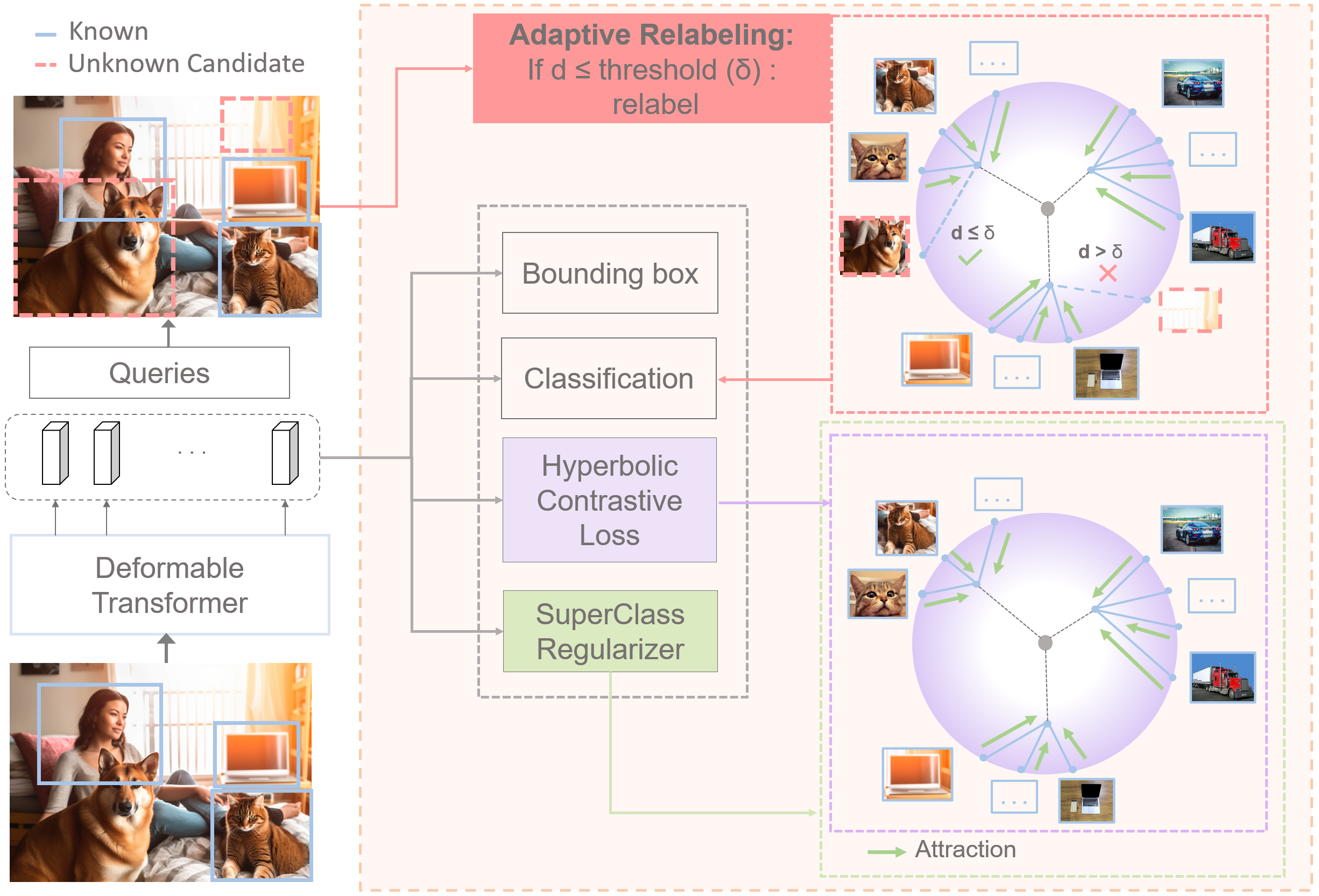}
    \end{center}
 
  \caption{\textbf{Overview of Hyp-OW}. Comprising three core components: the \textit{Hyperbolic Contrastive Loss} for representation learning at the class level; the \textit{SuperClass Regularizer}, for semantic relationships at the category level; and the \textit{Adaptive Relabeling} module, for unknown retrieval with the previously learned representation. If a distance $d$ between a candidate proposal and known items is lower than a certain threshold ($\delta$), the proposal is relabelled as unknown.}
  \label{fig:hypow_cartoon}
\end{figure*}

In this section, we provide a detailed explanation of each module of our proposed method. \method can be summarized by three main components (Figure~\ref{fig:hypow_cartoon}): a Hyperbolic Metric Distance learning, a SuperClass Regularizer, and an Adaptive Relabeling Scheme to detect unknowns.

\subsection{Metric Learning with Hyperbolic Distance} \label{sec:contrastive_learning}

We learn feature representation in the hyperbolic embedding space using a contrastive loss. The idea is to move closer features belonging to the same class $c$\footnote{By abuse of notation, we will also use $c$ for the class label since the meaning can be inferred from the context} while repelling them from features of different classes. Let's denote $\mb{z_{i}^{c}}$ any query $i$ matched with class $c \in \cK$. To facilitate readability, we will omit the index $c$ whenever there is no confusion about the class context.

Throughout training, we maintain a replay buffer $\cM$ where we store $m$ embedding features per class.
For every query element $\mb{z_{i}}$ of the incoming batch $\cB$, we sample $k=1$ element of the same class from the replay buffer $\cM$ denoted $\mb{z_{i^{+}}}$ (this element serves as the positive comparison) and consider the $2|\cB|-2$ remaining samples as the negative examples $\mb{z_{i^{-}}}$.

If we denote $\cA= \cB \cup \cM$ and define  a temperature $\tau_1$, the contrastive loss is then expressed as:

% Defining a temperature $\tau_1$, the contrastive loss is then expressed as:
\begin{align}
\mathscr{L}_{hyp}= - \displaystyle{\sum_{i \in \cA }}\log \frac{\exp(\frac{-d_{hyp}(\mb{z_{i}},\mb{z_{i^{+}}})}{\tau_1})}{\displaystyle{\sum_{i^{-} \in \cA \setminus \{i , i^{+}\}}} \exp(\frac{-d_{hyp}(\mb{z_{i}},\mb{z_{i^{-}}})}{\tau_1})}
\label{eq:contrastive_loss}
\end{align}
This loss aims at attracting representation of $\mb{z_{i}}$ closer to its positive counterpart $\mb{z_{i^{+}}}$ while repelling from the negative examples $\mb{z_{i^{-}}}$, $ i \in \cA$.

\subsection{SuperClass Regularization} \label{sec:superclass_regularizer}

Numerous real-world datasets inherently possess hierarchical structures, allowing classes to be categorized. For instance, dogs and cats fall under the broader category of "animals," while cars and trucks belong to the category of "vehicles." To harness this inherent hierarchy, we introduce a SuperClass Regularizer (we will use "SuperClass" and "category" interchangeably in this context). In contrast to Eq~\ref{eq:contrastive_loss}, our proposed regularization encourages grouping at the SuperClass level rather than the class level.

Let's denote $\cS_{p}$ as the set of class indexes within categories $p=1...P$ (we denote this set $\cP$). For instance, the category "vehicles" might encompass classes such as car, truck, bus, and so on... We approximate the category $p$ embedding by computing the Hyperbolic Average~\citep{hyperbolic_average} (dubbed $\textit{HypAve}$) of every embedding $\{ \mb{z_{i}^{c}} \}_{i \in \cM }$ of classes $c$ from the buffer $\cM$ within this category ($ c \in \cS_{p}$) that is:

\begin{align}
\mb{\overline{z}_{p}}=HypAve(\{ \mb{z_{i}^{c}} \}_{i \in \cM, c \in \cS_{p}})=\frac{\displaystyle{\sum_{i \in \cM,c \in \cS_{p}}} \gamma_{i}\mb{z_{i}^{c}}}{\displaystyle{\sum_{i \in \cM}} \gamma_{i}}
\label{eq:category_embedding}
\end{align}

where $\gamma_{i}=\frac{1}{\sqrt{1-c \lVert x_{i}  \rVert^{2} }}$ is the Lorentz factor. For each element $\mb{z_{i}^{c}}$ of a batch $\cB$, we extract its category embedding $\mb{\overline{z}_{p}}$  $(c \in \cS_{p})$ from the buffer $\cM$. Using a temperature parameter $\tau_2$, we formulate our SuperClass regularizer as follows:

\begin{align}
    \mathscr{L}_{reg}=\displaystyle{\sum_{i \in \cA, c \in \cS_{p}  }} -\log \frac{\exp(\frac{-d_{hyp}(\mb{z_{i}^{c}},\mb{\overline{z}_{p}})}{\tau_2})}{\displaystyle{\sum_{k \neq p }} \exp(\frac{-d_{hyp}(\mb{z_{i}^{c}},\mb{\overline{z}_{k}})}{\tau_2})}
    \label{eq:family_regularizer}
\end{align}

This loss encourages the features $\mathbf{z}_i^c$ of each class $c$ to be closer to its corresponding category embedding $\mathbf{\overline{z}}_p$, while simultaneously pushing it away from embeddings of other categories $\mathbf{\overline{z}}_k$, $k \neq p$. In essence, it fosters the grouping of similar items at the SuperClass/category level rather than the individual class level.

\subsection{Adaptive Relabeling of Unknowns with Hyperbolic Distance}
\label{sec:adaptive_relabelling}
We introduce our Adaptive Relabeling module, which \textit{dynamically adapts} to the batch statistics to effectively detect unknowns. 
We can summarize this procedure in three steps: \textcolor{Fuchsia}{a) the hyperbolic mean, or \textit{centroid}, is calculated for each class in $\mathcal{M}$,} \textcolor{teal}{b) for all known items in an image, we compute the distance to each centroid, the greatest distance is labeled as $\delta_{B}$}, \textcolor{YellowOrange}{c) for every unmatched bounding box, we calculate its distance to each centroid: if less than $\delta_{B}$, it is relabeled as 'unknown' otherwise it is considered as a background.}

\textcolor{Fuchsia}{a)} We define $\mb{\underline{z}_{c}}$ \footnote{We differentiate from $\mb{\overline{z}_{p}}$ with an underline to distinguish Hyperbolic Average of class and category} the hyperbolic average of class $c$ computed from the buffer $\cM$ as: 

$HypAve(\{ \mb{z_{i}^{c}} \}_{i \in \cM})=\frac{\displaystyle{\sum_{i \in \cM}} \gamma_{i}\mb{z_{i}^{c}}}{\displaystyle{\sum_{i \in \cM}} \gamma_{i}}  $ which can be seen as the centroid of each class $c$ in the hyperbolic embedding space.

\textcolor{teal}{b)} We now use the matched queries to define: $\delta_{\cB}=\displaystyle{\max_{i \in \cB, c \in \cK}} \; d_{hyp}(\mb{z_{i}},\mb{\underline{z}_{c}})$. 
In essence, $\delta_{\cB}$ signifies the greatest distance from any known items in the batch $\cB$ to all centroid $\mb{\underline{z}_{c}}, c\in \cK$ in the replay buffer $\cM$. 

 \textcolor{YellowOrange}{c)} This threshold is then utilized to relabel any unmatched query $\mb{z_{u}}$ as unknown if:
\begin{align}
\displaystyle{\min_{c \in \cK}} \; d_{hyp}(\mb{z_{u}},\mb{\underline{z}_{c}}) \leq \delta_{\cB}
\label{eq:adaptive_relabeling}
\end{align}

\paragraph{Overall loss}
All the aforementioned losses are finally optimized together as:
\begin{align}
    \cL= \cL_{cls}+ \cL_{bbox} + \alpha \cL_{hyp}+ \beta \mathcal{L}_{reg}  
\end{align}
Where $\alpha,\beta \geq 0$ are coefficient controlling respectively the Hyperbolic and regularizer importance.

%% file: 5-experiments.tex
\section{Experiments}
In this section, we start with describing our experimental setup. We then present comparative results against benchmark baselines, followed by in-depth ablation analysis of each component of \method. Due to space limitations, we will defer detailed information to the Supplementary Material.

\subsection{Experimental Setup}

\begin{table*}[t]
\centering
\resizebox{0.9\textwidth}{!}{
\begin{tabular}
{c|l|cc||cc||cc||c} \hline
 \multirow{2}{*}{\rotatebox[origin=c]{90}{{\tiny\textbf{Regime}}}} &  &\multicolumn{2}{c}{\textbf{Task 1}} & \multicolumn{2}{c}{\textbf{Task 2}}& \multicolumn{2}{c}{\textbf{Task 3}} & \multicolumn{1}{c}{\textbf{Task 4}} \\ \cline{3-9}   
  & Methods   &  U-Recall     ($\uparrow$)       & mAP  ($\uparrow$)  &     U-Recall  ($\uparrow$)        &  mAP   ($\uparrow$)         &        U-Recall ($\uparrow$)  &  mAP ($\uparrow$)  & mAP ($\uparrow$)  \\  \hline
\multirow{4}{*}{\rotatebox[origin=c]{90}{\textcolor{Periwinkle}{\textbf{Low}}}}
&  ORE - EBUI            & 1.5  &  61.4 &   3.9  &  40.6 &   3.6  &    33.7  &    31.8 \\
&  OW-DETR              & 5.7 &71.5 & 6.2 & 43.8 & 6.9 & 38.5 & 33.1 \\
& PROB                               &    17.6 & \textbf{73.4}  & 22.3 & 50.4 & 24.8  & 42.0  &  39.9 \\
 & \method (Ours)                                   & \textbf{23.9}   &   72.7   & \textbf{23.3}    & \textbf{50.6}   & \textbf{25.4}    & \textbf{46.2}    & \textbf{44.8}   \\  
 \hline
 &  $\Delta$(Rel. Difference)                                   &  \textcolor{Green}{\textbf{+6.3}}  &  $\leq 1.0$ & \textcolor{Green}{\textbf{+1.0}}   & $\leq 1.0$   & $\leq 1.0$    &  \textcolor{Green}{\textbf{+4.2}}    &   \textcolor{Green}{\textbf{+4.9}} \\   \hline
\multirow{7}{*}{\rotatebox[origin=c]{90}{\textcolor{LimeGreen}{\textbf{Medium}}}} & ORE - EBUI    & 4.9   & 56.0   &  2.9   &  39.4  &   3.9  & 29.7  & 25.3   \\ 
&  UC-OWOD                     & 2.4   & 50.7   &  3.4   &  8.7   & 16.3 & 24.6 & 23.2   \\ 
&   OCPL                         & 8.26  & 56.6   &  7.65 &   39.1  &  11.9 & 30.7 & 26.7   \\ 
 & 2B-OCD                      & 12.1  & 56.4   &  9.4    &  38.5  & 11.6 & 29.2 &  25.8   \\
  &OW-DETR      & 7.5   & 59.2   &  6.2    &  42.9  &   5.7  & 30.8  & 27.8   \\ 
 &  PROB          & 19.4  & \textbf{59.5}   & 17.4  &  44.0  & 19.6 & 36.0 & 31.5      \\ 
 &  \method (Ours)                             & \textbf{23.5}   & 59.4    & \textbf{20.6}   & \textbf{44.4}  &  \textbf{26.3}  &  \textbf{36.8} &    \textbf{33.6}    \\   \hline
  &  $\Delta$(Rel. Difference)                              &  \textcolor{Green}{\textbf{+4.1}}  & $\leq 1.0$    & \textcolor{Green}{\textbf{+3.2}}   &  $\leq 1.0$  &  \textcolor{Green}{\textbf{+6.7}}  &  $\leq 1.0$  &     \textcolor{Green}{\textbf{+2.1}}    \\  \hline 
\parbox[t]{3mm}{\multirow{3}{*}{\rotatebox[origin=c]{90}{\textcolor{Red}{\textbf{High}}}}} & OW-DETR              & 7.0 &  47.3  & 11.0   &  38.6  &  8.8 &  38.3 &  38.2 \\
& PROB                                  & 29.4 & 49.6  & 43.9   & 42.9   & 52.7   & 41.3 &41.0\\
 & \method (Ours)                                  &  \textbf{34.9}  & \textbf{49.9}    & \textbf{47.5}  &  \textbf{45.5}  & \textbf{55.2}   &  \textbf{44.3} \ &   \textbf{43.9}\\   \hline
& $\Delta$(Rel. Difference)                            &   \textcolor{Green}{\textbf{+5.5}} & $\leq 1.0$   &  \textcolor{Green}{\textbf{+3.6}} &   \textcolor{Green}{\textbf{+2.6}}  &   \textcolor{Green}{\textbf{+2.5}} &   \textcolor{Green}{\textbf{+3.0}} &    \textcolor{Green}{\textbf{+2.9}}\\   \hline 
% \hhline{|=|=|=|=|=|=|=|=|=|}
\end{tabular}
}
\caption{\textbf{State-of-the-art comparison on the three splits for unknown detection (U-Recall) and known accuracy (mAP).} \method improves significantly the unknown detection (U-Recall) for the medium and high regime and known detection (mAP) for the low regime. Task 4 does not have U-Recall since all 80 classes are known at this stage.}
\label{tab:final_mini_results}
\end{table*}

\paragraph{Implementation Details}
We use Deformable DETR \citep{deformable_transformer} pretrained in a self-supervised manner (DINO \cite{caron2021emerging}) on Resnet-50 \citep{resnet50} as our backbone. The number of deformable transformer encoder and decoder layers are set to $6$. The number of queries is set to $Q=100$ with a dimension $d=256$. During inference time, the top-100 high scoring queries per image are used for evaluation. For our method, We used $c=0.1$, $\tau_1=0.2$, $\tau_2=0.4$. For the set $S_p$ that defines the composition of each category (SuperClass), we adhere to the grouping used in MS-COCO dataset~\citep{mscoco}. All used hyperparameters can be found in the Supplementary.

\paragraph{Metrics and Baselines} 
Following the current metrics used for OWOD, we utilize the mean average precision (mAP) for known items, while U-Recall serves as the primary metric to quantify the quality of unknown detection for each method~\citep{owdetr,two_branch,prob,maaz2022class,owod_discriminative_prototypes}. Additional metric is discussed in Table~\ref{tab:unknown_confusion_main}. We consider the following baselines from literature: OW-DETR~\citep{owdetr} and PROB~\citep{prob}. While we included Faster R-CNN methods as informative references (ORE-EBUI~\citep{owod}, UC-OWOD~\citep{uc_owod}, OCPL~\citep{owod_discriminative_prototypes}, 2B-OCD~\citep{two_branch}), our primary emphasis is on comparing against deformable Transformer-based methods to ensure a fair assessment following the evaluation procedure of PROB.

\paragraph{Datasets}
We consider two benchmarks from the literature: the OWOD Split \cite{owod} and the OWDETR Split \cite{owdetr}. While the latter (OWDETR Split) strictly separates SuperClasses across tasks the first (OWOD) has mild semantic overlap between knowns and unknowns across tasks (See Supplementary Material). To closely mimic real-world scenarios, we consider a \textit{Hierarchical Split} which ensures that each task includes at least one class from each category\footnote{While there are various ways to distribute the classes across tasks, our primary intention here is to introduce a third scenario for assessing all methods.}. Each dataset is defined by four tasks $t=1,2,3,4$, containing 20 labelled classes each, for a total of 80 classes. When task $t$ starts, only the label of classes belonging to that task are revealed. For instance, task $1$ only contains labels of classes from 0 to 19, while task $2$ only contains labels of classes from 21 to 39, and so on. Composition of each dataset can be found in the Supplementary.

\begin{figure}[h!]
  \begin{center}
    \includegraphics[width=0.43\textwidth]{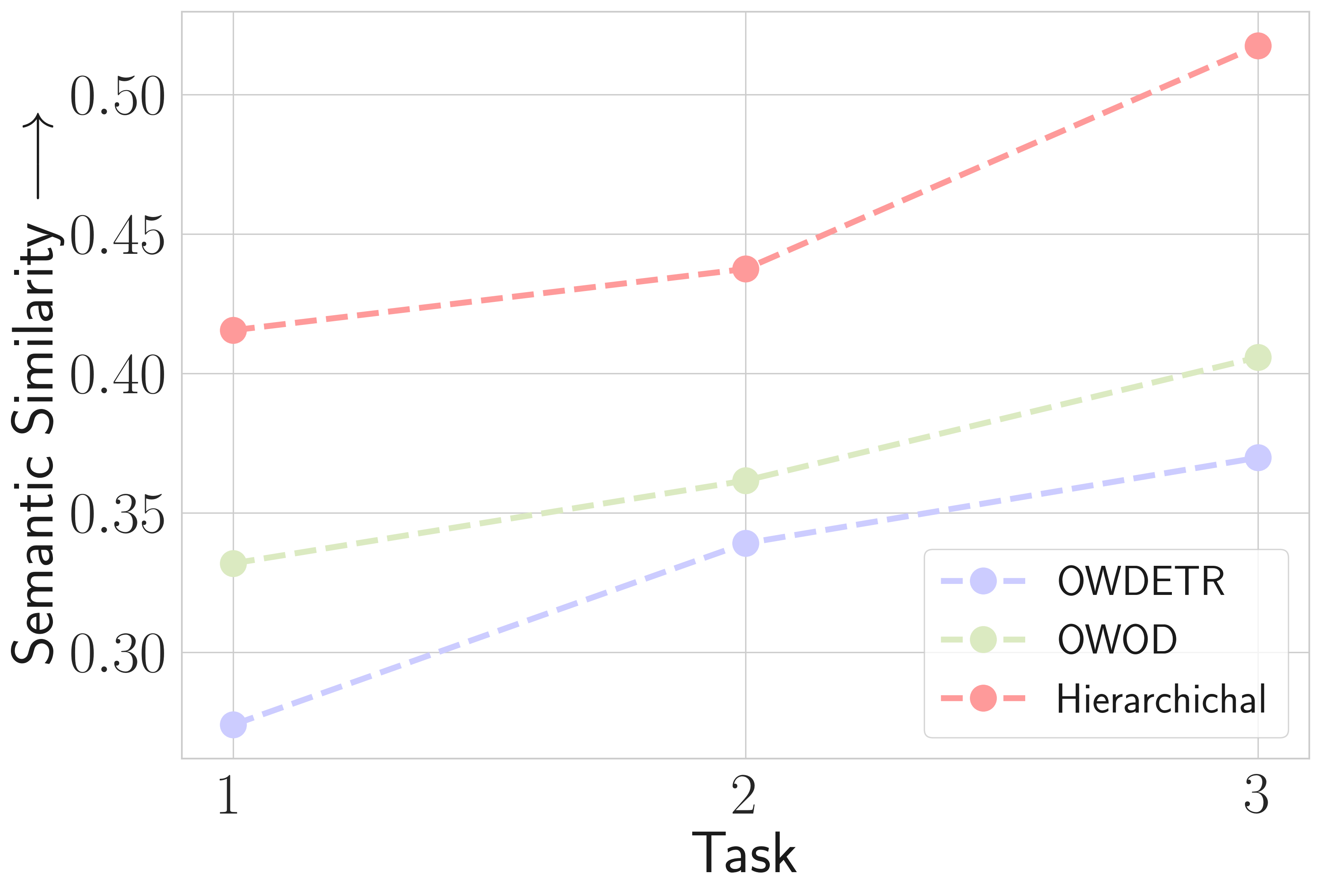}
  \end{center}
  \caption{\textbf{Semantic Similarity between knowns and unknowns across tasks for each Split.}}
  \label{fig:dataset_similarity_main}
\end{figure}

\paragraph{Dataset Structure} \label{paragraph:dataset_similarity}
To better understand the structure of each dataset, we define a semantic similarity measure using GloVe's embedding~\cite{pennington-etal-2014-glove}. Denoting $\mb{\omega_{c}}, c \in \cK$ ( respectively $\mb{\omega_{k}}, c \in \cU$ ) the embedding of known (unknown) classes, the semantic overlap between knowns and unknowns for task $t \in [1,T-1]$ is:

\begin{align}
S_{t}=\frac{1}{|\cU^{t}|} \displaystyle{\sum_{k \in \cU^{t}}  \max_{c \in \cK^{t}}} 
\frac{ < \mb{\omega^{c}},\mb{\omega^{k} >}}  {\lVert \mb{\omega^{c}}  \rVert \cdot \lVert \mb{\omega^{k}}  \rVert}  
\label{eq:semantic_measure_main}
\end{align}
Higher values indicate larger overlap between the knowns and unknowns. Figure~\ref{fig:dataset_similarity_main} effectively quantifies the level of hierarchical structure inherent in each dataset, providing a clear insight  into the composition of each split. This framework now offers us a consistent basis to evaluate each method across different hierarchical scenarios: \textcolor{Periwinkle}{Low regime} (OW-DETR Split), \textcolor{LimeGreen}{Medium regime} (OWOD Split) and \textcolor{Red}{High regime} (Hierarchical Split). This metric is increasing as the number of knowns grows throughout the training

\subsection{Benchmark Results}

\paragraph{Unknown Detection (U-Recall)}
Table~\ref{tab:final_mini_results} shows the high performance gain of \method over PROB on \mediumregime and \highregime  of $3\%$ on average (The row $\Delta$ indicates relative performance with respect to the second best algorithm). This highlights the utility of learning hierarchical structured representations and retrieving unknowns based on their similarity with known objects, as opposed to PROB, which learns a single mean representation for all objects.
For the \lowregime our method is performing on-par with PROB except for task 1 which shows a surprising improvement of $6$ points. Overall, \method demonstrates consistent and strong results on the three benchamrks.

\paragraph{Known Accuracy (mAP)}
\method outperforms baseline benchmarks across all tasks in the Hierarchical Split and notably enhances performance for the last two tasks in the OW-DETR Split. This success can be attributed to the learned structural hierarchy, which groups classes of the same category (illustrated in t-SNE Figure~\ref{fig:tse_mini}). Moreover, our method exhibits robust performance even in the low regime. This can be attributed to the inherent presence of bounding box overlaps in object detection tasks, enabling the model to learn about co-occurring objects (refer to Supplementary Material).

\begin{table}[b]
\centering
\resizebox{0.45\textwidth}{!}{
\begin{tabular}
{l|cc} \hline
       &  U-Recall ($\uparrow$)  & mAP($\uparrow$)   \\ \hline
$c=0.0$ (Cosine Dist.)                 &  32.8  & 49.0     \\   \hline
   $c=0.1$ (\method)                 &  \textbf{34.9}  & \textbf{49.9}     \\ \hline
   $c=0.2$               &  33.3  &  49.5    \\ \hline
  $c=0.5$                 &  32.3 & 49.8     \\ \hline
\end{tabular}
}
\caption{\textbf{Impact of curvature coefficient $c$ for Hierarchical Split Task 1}($\uparrow$ indicates larger values are better and $\downarrow$ indicates smaller values are better).}
\label{tab:ablation_curvature}
\end{table}

\begin{figure}[h!]
\includegraphics[width=1.0\linewidth]{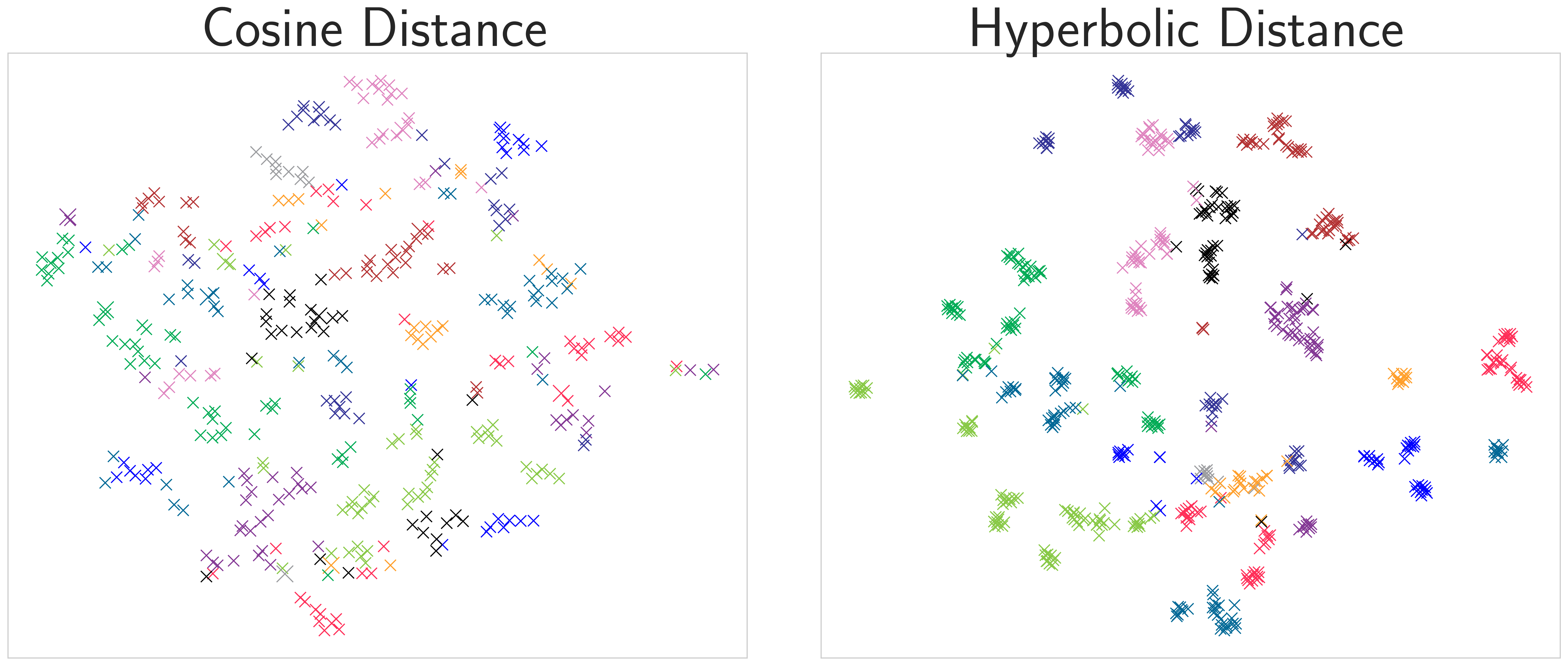}
  \caption{\textbf{t-SNE plot of the learned class representations} Hyperbolic Distance tends to learns a better hierarchical structure than Cosine Distance.}
  \label{fig:tse_mini_cos_hyp}
\end{figure}

\begin{table*}[ht!]
\resizebox{0.95\textwidth}{!}{
 \hspace{10mm}
\begin{tabular}
{l|cc||cc||cc||c}\hline
  & \multicolumn{2}{c}{\textbf{Task 1}} & \multicolumn{2}{c}{\textbf{Task 2}}& \multicolumn{2}{c}{\textbf{Task 3}} & \multicolumn{1}{c}{\textbf{Task 4}} \\ \cline{2-8} 
      &  U-Recall ($\uparrow$) & mAP ($\uparrow$)  &  U-Recall ($\uparrow$) &  mAP ($\uparrow$)  & U-Recall ($\uparrow$)&  mAP ($\uparrow$) &mAP ($\uparrow$)\\ \hline
 \method (Ours)                 &  \textbf{34.9}  & 49.9    & 47.5     & \textbf{45.5}   &  55.2   &  \textbf{44.3}  &   \textbf{43.9}  \\ 
 w/  Cosine Distance (c=0)         & 32.8     & 49.0   &  46.4  & 45.4 & \textbf{55.4} &   43.2   &  43.1\\
w/o SuperClass Regularizer        & 32.0   & \textbf{50.0} &  47.1 & 45.1 & 52.9   &  43.7 &  43.5 \\
 w/o  Adaptive Relabeling          & 34.7 & 41.2     & \textbf{47.6}    & 38.9      & 54.1   &  36.5  l & 36.1    \\ \hline
\end{tabular}
}
\caption{\textbf{Impact of each component of \method on Hierarchical Split}. We observe that the Relabeling module (third line) significantly reduces mAP while maintaining U-Recall. On the other hand, the SuperClass Regularizer and Cosine Distance have primarily an impact on unknown detection. Task 4 does not have U-Recall since all 80 classes are known at this stage. }
\label{tab:main_ablation}
\end{table*}

\subsection{Ablation Analysis}
We now aim to gain an in-depth understanding of \method by systematically removing each component one by one to assess its direct impact (illustrated quantitatively in Table~\ref{tab:main_ablation} for Hierarchical Split and in the Supplementary for OWOD-Split). Additionally, we perform a quantitative analysis of the unknown confusion. Qualitative visualizations can be found in the supplementary.

\begin{figure*}[ht!]

    \includegraphics[width=1.0\linewidth]{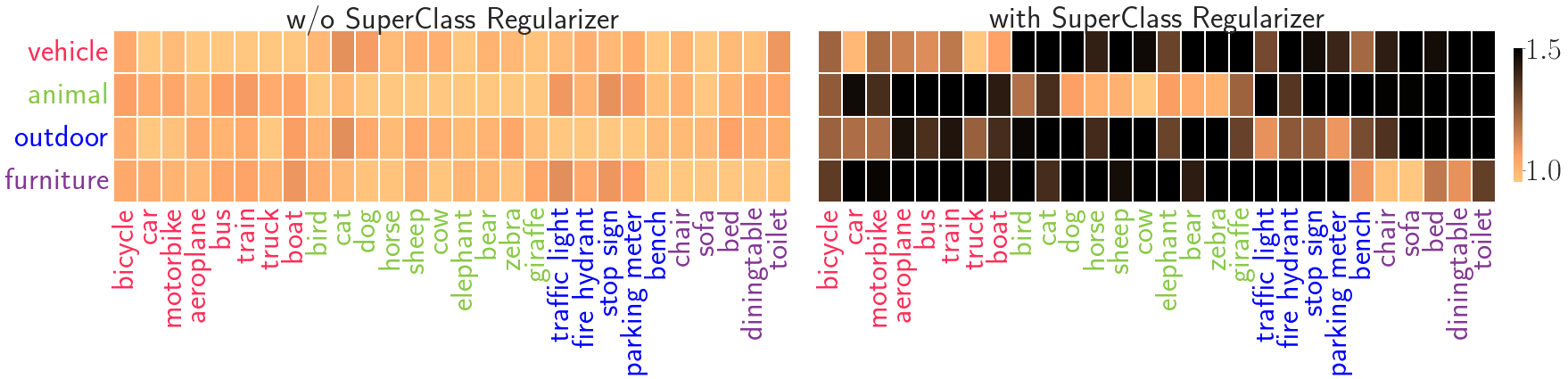}

  \caption{\textbf{Hyperbolic Category - Class Distance Heatmap.} The SuperClass Regularizer (right) effectively separates different categories (left), as indicated by the increased distance between each animal class (bottom) and the vehicle, outdoor, and furniture categories (darker colors). Without this regularizer (left), category inter-distance are much smaller (lighter color intensity).}
  \label{fig:ablation_superclass_regularizer_mini}
  \vspace{-5mm}
\end{figure*}

\paragraph{Curvature Coefficient $c$:} We assess the impact of different hyperbolic distances alongside the cosine distance $c=0$ (which is also linked with the Euclidian distance for normalized vectors, See Supplementary) in Table~\ref{tab:ablation_curvature}. While substituting hyperbolic distance with cosine distance ($c=0$) negatively affects both U-Recall and mAP, higher $c$ degrades mainly the U-Recall. The hyperbolic embedding space is more suitable to learn data with latent hierarchical structure (see t-SNE plot Figure~\ref{fig:tse_mini_cos_hyp}).

\paragraph{Adaptive Relabeling:}

The relabeling of unmatched bounding boxes as unknowns is governed by Eq~\ref{eq:adaptive_relabeling}. To evaluate its impact, we constrast it with an alternative technique used by PROB~\citep{prob}, where all unmatched queries are classified as unknowns. Results are shown in Table~\ref{tab:main_ablation} fourth row.  Although the decrease in U-Recall is marginal, a noteworthy reduction in known accuracy (mAP) is observed. This decline can be attributed to the over-prediction of patches as unknowns, which results in misclassification of known objects. More detailed figures in the Supplementary show its efficacy where we notice that unknowns belonging to the same category as knowns exhibit lower Hyperbolic Distance, manifested as lighter colors.

\paragraph{SuperClass Regularizer:} By setting $\beta=0$ (Table~\ref{tab:main_ablation}: third row), we no longer enforce the grouping of items at the category level (compare t-SNE plots in Figure~\ref{fig:tse_mini}). We then observe a reduction in U-Recall of $2.9$, $0.4$, and $2.3$ points respectively. Heatmap Figure~\ref{fig:ablation_superclass_regularizer_mini} illustrates the hyperbolic distance from each class to every category's embedding (computed using Eq~\ref{eq:category_embedding}) with lighter colors indicating smaller distances. With our regularizer (right plot), we observe a wider range of values spanning from $0.7$ to $2.30$, compared to a smaller range of $0.78$ to $1.2$ without the regularizer. This emphasizes the impact of our regularizer, which effectively separates classes from distinct categories (depicted by the darker color in the right plot) while simultaneously bringing classes from similar category closer together. A more detailed plot can be found in the Supplementary.

\paragraph{Unknowns Confusion:}
We measure the A-OSE metric introduced by ~\citet{owod} (defined in the Supplementary Material) which quantifies the number of unknowns misclassified as knowns (lower is better). Table~\ref{tab:unknown_confusion_main} showcases the results for Hierarchichal Split (see in Supplementary Material for OWOD Split). Comparing with PROB, \method exhibits significantly fewer misclassifications across different tasks, with at least a twofold reduction for the first two tasks and approximately $20\%$ less for the third task.

\begin{table}[h!]
\centering
% \hspace{-14mm}
\resizebox{0.45\textwidth}{!}{
\begin{tabular}
{l|c||c||c} \hline
      &\multicolumn{1}{c}{\textbf{Task 1}}  &\multicolumn{1}{c}{\textbf{Task 2}} &\multicolumn{1}{c}{\textbf{Task 3}} \\ \cline{2-4} 
       &  A-OSE($\downarrow$)  &   A-OSE($\downarrow$) &   A-OSE($\downarrow$)  \\ \hline
OW-DETR                  & 42,540    & 26,527 &  20,034\\   \hline
 PROB              &  14,962    &  8,929 &    5,387 \\ \hline
  \method (Ours)             &    \textbf{7,420}    & \textbf{3,849} &  \textbf{4,611}  \\ \hline 
  % \hhline{|=|=|=|=|} 
  % CAT                 &  8.5 & 44,701     & 12.0 & 24,748& 13.7  &  17,394 \\ \hline
\end{tabular}
}
\caption{\textbf{A-OSE metric on Hierarchical Split.} \method exhibits a lower rate of unknowns misclassifition as knowns compared to other baselines.}
\label{tab:unknown_confusion_main}
\vspace{-5mm}
\end{table}

%% file: 6-conclusion.tex
\section{Conclusion}
The Open World Object Detection framework presents a challenging and promising setting, encompassing crucial aspects such as lifelong learning and unknown detection. In our work, we have emphasized the lack of a clear definition of unknowns and the need for a hierarchical or semantic relationship between known and unknown classes. This led us to propose \method that focuses on learning and modeling the structural hierarchy within the dataset, which is then utilized for unknowns retrieval. Extensive experiments demonstrate significant improvement of \method for both known and unknown detection (up to 6 percent) particularly in the presence of inherent hierarchy between classes. Future directions include leveraging knowledge from pretrained vision language models to detect desired unknowns~\citep{zohar2023open}. We also hope that our hierarchical structural learning paradigm benefits adjacent fields such as OOD detection~\citep{behpour2023gradorth}, data pre-selection~\citep{li2023updp} or instance segmentation~\citep{instance_seg}.

%% file: appendix.tex
\onecolumn

\section{Supplementary Material}
In the next secions we respectively cover the following topics:
\begin{itemize}
    \item Dataset composition for each split and introduction of our semantic similarity measure
    \item Experimental details
    \item Qualitative visualization demonstrating the effectiveness of Hyp-OW in comparison to PROB and individual components of Hyp-OW
    \item Comprehensive benchmark metrics results and visual plots (heatmaps) to illustrate our findings
    \item Ablation analysis discussion accompanied with their qualitative plots
\end{itemize}

\newpage

\section{Dataset Information} \label{app:dataset_information}
\subsection{Dataset Composition}
In this section, we provide the composition of each dataset split and explain the main differences between them. The splits are made with images taken from the PASCAL-VOC2007/2012~\citep{pascalvoc} and MS-COCO dataset~\citep{mscoco}, containing a total of 80 classes grouped into 12 catogories (SuperClass): \textcolor{red}{vehicle}, \textcolor{blue}{outdoor}, \textcolor{LimeGreen}{animal}, \textcolor{YellowOrange}{accessory}, \textcolor{MidnightBlue}{sports}, \textcolor{Thistle}{kitchen}, \textcolor{Green}{food}, \textcolor{Fuchsia}{furniture}, appliance, \textcolor{Mahogany}{electronic}, \textcolor{BlueViolet}{indoor}, and \textcolor{Gray}{person}. We categorize the different datasets into three regimes based on a semantic similarity metric (defined below), which quantifies the overlap between known and unknown items:
\begin{itemize}
\item \highregime: Hierarchical Split
\item \mediumregime: OWOD Split
\item \lowregime: OWDETR Split

\end{itemize}
The low regime implies no semantic similarity between classes of different tasks. For instance, OWDETR Split's (Table~\ref{tab:owdetr_split}) classes of task 1 are composed of vehicle, animal, and person, which do not appear in tasks 2, 3, or 4. On the other hand, high regime shows high similarity between classes across all tasks. Each task of the Hierarchical split (Table~\ref{tab:hierarchical_split}) contains at least one class from each of the 12 categories (Table~\ref{tab:stats_hierarchical} shows the statistics of the split). The OWOD Split is a trade-off between the aforementioned datasets.

\begin{table}[h!]
\centering

\begin{tabular}{c|c|c|c}
\hline
\multicolumn{4}{c}{ \highregime: \textbf{hierarchical Split}} \\ \hline
Task 1  & Task 2 & Task 3 & Task 4 \\ \hline
\textcolor{OrangeRed}{bicycle}     & \textcolor{Gray}{person}  &     \textcolor{red}{bus} & \textcolor{red}{truck} \\ \hline
\textcolor{OrangeRed}{car}     &       \textcolor{red}{motorbike}  &  \textcolor{red}{train} &  \textcolor{red}{boat}\\ \hline
\textcolor{blue}{traffic light}     &  \textcolor{red}{aeroplane}  &  \textcolor{blue}{parking meter}&  \textcolor{blue}{bench}\\ \hline
\textcolor{blue}{fire hydrant}     &\textcolor{blue}{stop sign}  & \textcolor{LimeGreen}{cow} &  \textcolor{LimeGreen}{zebra} \\ \hline
\textcolor{LimeGreen}{bird}     & \textcolor{LimeGreen}{horse}  & \textcolor{LimeGreen}{elephant} &  \textcolor{LimeGreen}{giraffe}\\ \hline
\textcolor{LimeGreen}{cat}     & \textcolor{LimeGreen}{sheep} &  \textcolor{LimeGreen}{bear} & \textcolor{YellowOrange}{tie} \\ \hline
\textcolor{LimeGreen}{dog}     & \textcolor{YellowOrange}{umbrella} & \textcolor{YellowOrange}{handbag} & \textcolor{YellowOrange}{suitcase} \\ \hline
\textcolor{YellowOrange}{backpack}     & \textcolor{MidnightBlue}{snowboard} & \textcolor{MidnightBlue}{kite} & \textcolor{MidnightBlue}{skateboard} \\ \hline 
\textcolor{MidnightBlue}{frisbee}     & \textcolor{MidnightBlue}{sports ball} & \textcolor{MidnightBlue}{baseball bat} &  \textcolor{MidnightBlue}{surfboard}\\ \hline
\textcolor{MidnightBlue}{skis}     & \textcolor{Thistle}{cup}  & \textcolor{MidnightBlue}{baseball glove} &  \textcolor{MidnightBlue}{tennis racket}\\ \hline
\textcolor{Thistle}{bottle}     & \textcolor{Green}{sandwich} & \textcolor{Thistle}{fork} &  \textcolor{Thistle}{spoon} \\ \hline
\textcolor{Thistle}{wine glass}     & \textcolor{Green}{orange} & \textcolor{Thistle}{knife}&  \textcolor{Thistle}{bowl}  \\ \hline
\textcolor{Green}{banana}     & \textcolor{Green}{broccoli} & \textcolor{Green}{carrot} & \textcolor{Green}{pizza} \\ \hline
\textcolor{Green}{apple}     & \textcolor{Fuchsia}{pottedplant} & \textcolor{Green}{hot dog} &  \textcolor{Green}{donut}\\ \hline
\textcolor{Fuchsia}{chair}     & \textcolor{Fuchsia}{bed} &  \textcolor{Fuchsia}{diningtable} &  \textcolor{Green}{cake} \\ \hline  
\textcolor{Fuchsia}{sofa}     & \textcolor{Mahogany}{laptop} & \textcolor{Mahogany}{remote} &  \textcolor{Fuchsia}{toilet} \\ \hline
\textcolor{Mahogany}{tvmonitor}     & \textcolor{Mahogany}{mouse} & \textcolor{Mahogany}{keyboard} &   \textcolor{Mahogany}{cell phone} \\ \hline
microwave     & toaster & sink &  refrigerator \\ \hline
oven    & \textcolor{BlueViolet}{clock} & \textcolor{BlueViolet}{scissors} &  \textcolor{BlueViolet}{hair drier} \\ \hline
\textcolor{BlueViolet}{book}     & \textcolor{BlueViolet}{vase} &  \textcolor{BlueViolet}{teddy bear} & \textcolor{BlueViolet}{toothbrush} \\ \hline
\end{tabular}

\centering

\vspace{+5mm}
\caption{\textbf{Composition of hierarchical Split.} Each task contains at least one class of each category.}
\label{tab:hierarchical_split}
\end{table}

\begin{table}[h!]
\centering
% \hspace{-14mm}
\resizebox{0.45\textwidth}{!}{
\begin{tabular}
{l|c||c||c||c} \hline
      &\multicolumn{1}{c}{\textbf{Task 1}}  &\multicolumn{1}{c}{\textbf{Task 2}} &\multicolumn{1}{c}{\textbf{Task 3}}  & \multicolumn{1}{c}{\textbf{Task 4}}\\ \cline{1-5} 
$\#$ training images          & 28,677    & 38,552 &  27,147  & 34,136\\   \hline
 $\#$ test images             &  4,952    &  4,952 & 4,952  & 4,952  \\ \hline
 $\#$ unknowns             &  26,523    &  10,476 &  5,441 & N/A   \\ \hline
\end{tabular}
}
\caption{\textbf{Statistics of Hierarchical Split.}}
\label{tab:stats_hierarchical}
\end{table}

% \subsection{OWOD Split}
\begin{table}[h!]
\centering

\begin{tabular}{c|c|c|c}
\hline
\multicolumn{4}{c}{\mediumregime: \textbf{OWOD Split}} \\ \hline
Task 1  & Task 2 & Task 3 & Task 4 \\ \hline
\textcolor{OrangeRed}{aeroplane}     & \textcolor{OrangeRed}{truck}  &     \textcolor{MidnightBlue}{frisbee} & \textcolor{Fuchsia}{bed} \\ \hline
\textcolor{OrangeRed}{bicycle}     &       \textcolor{blue}{traffic light}  &  \textcolor{MidnightBlue}{skis} &  \textcolor{Fuchsia}{toilet}\\ \hline
\textcolor{LimeGreen}{bird}     &  \textcolor{blue}{fire hydrant}  &  \textcolor{MidnightBlue}{snowboard}&  \textcolor{Mahogany}{laptop}\\ \hline
\textcolor{OrangeRed}{boat}     &\textcolor{blue}{stop sign}  & \textcolor{MidnightBlue}{sports ball} &  \textcolor{Mahogany}{mouse} \\ \hline
\textcolor{Thistle}{bottle}     & \textcolor{blue}{parking meter}  & \textcolor{MidnightBlue}{kite} &  \textcolor{Mahogany}{remote}\\ \hline
\textcolor{OrangeRed}{bus}     & \textcolor{blue}{bench} &  \textcolor{MidnightBlue}{baseball bat} &  \textcolor{Mahogany}{keyboard} \\ \hline
\textcolor{OrangeRed}{car}     & \textcolor{LimeGreen}{elephant} & \textcolor{MidnightBlue}{baseball glove} & \textcolor{Mahogany}{cell phone} \\ \hline
\textcolor{LimeGreen}{cat}     & \textcolor{LimeGreen}{bear} & \textcolor{MidnightBlue}{skateboard} & \textcolor{BlueViolet}{book} \\ \hline 
\textcolor{Fuchsia}{chair}     & \textcolor{LimeGreen}{zebra} & \textcolor{MidnightBlue}{surfboard} &  \textcolor{BlueViolet}{clock}\\ \hline
\textcolor{LimeGreen}{cow}     & \textcolor{LimeGreen}{giraffe}  & \textcolor{MidnightBlue}{tennis racket} &  \textcolor{BlueViolet}{vase}\\ \hline
\textcolor{Fuchsia}{diningtable}     & \textcolor{YellowOrange}{backpack} & \textcolor{Green}{banana} &  \textcolor{BlueViolet}{scissors} \\ \hline
\textcolor{LimeGreen}{dog}     & \textcolor{YellowOrange}{umbrella} & \textcolor{Green}{apple}&  \textcolor{BlueViolet}{teddy bear}  \\ \hline
\textcolor{LimeGreen}{horse}     & \textcolor{YellowOrange}{handbag} & \textcolor{Green}{sandwhich} & \textcolor{BlueViolet}{hair drier} \\ \hline
\textcolor{OrangeRed}{motorbike}     & \textcolor{YellowOrange}{tie}  & \textcolor{Green}{hot dog} &  \textcolor{BlueViolet}{toothbrush}\\ \hline
\textcolor{Gray}{person}     & \textcolor{YellowOrange}{suitcase} &  \textcolor{Green}{broccoli} &  \textcolor{Thistle}{wine glass} \\ \hline  
\textcolor{Fuchsia}{pottedplant}     & microwave & \textcolor{Green}{carrot} &  \textcolor{Thistle}{cup} \\ \hline
\textcolor{LimeGreen}{sheep}     & oven & \textcolor{Green}{hot dog} &   \textcolor{Thistle}{fork} \\ \hline
\textcolor{Fuchsia}{sofa}     & toaster & \textcolor{Green}{pizza} &  \textcolor{Thistle}{knife} \\ \hline
\textcolor{OrangeRed}{train}     & sink & \textcolor{Green}{donut} &  \textcolor{Thistle}{spoon} \\ \hline
\textcolor{Mahogany}{tvmonitor}     & refrigerator &  \textcolor{Green}{cake} & \textcolor{Thistle}{bowl} \\ \hline
\end{tabular}

\centering

\vspace{+5mm}
\caption{\textbf{Composition of OWOD Split.} There is a mild overlap of categories between each task.}
\label{tab:owod_split}
\end{table}

\begin{table}[h!]
\centering

\begin{tabular}{c|c|c|c}
\hline
\multicolumn{4}{c}{  \lowregime: \textbf{OWDETR Split}} \\ \hline
Task 1  & Task 2 & Task 3 & Task 4 \\ \hline
\textcolor{OrangeRed}{aeroplane}     & \textcolor{blue}{traffic light}           &     \textcolor{MidnightBlue}{frisbee}    & \textcolor{Mahogany}{laptop} \\ \hline
\textcolor{OrangeRed}{bicycle}     &       \textcolor{blue}{fire hydrant}           &  \textcolor{MidnightBlue}{skis}       &  \textcolor{Mahogany}{mouse}\\ \hline
\textcolor{LimeGreen}{bird}     &  \textcolor{blue}{stop sign}                   &  \textcolor{MidnightBlue}{snowboard}    &  \textcolor{Mahogany}{remote}\\ \hline
\textcolor{OrangeRed}{boat}     &\textcolor{blue}{parking meter}                & \textcolor{MidnightBlue}{sports ball}      &  \textcolor{Mahogany}{keyboard} \\ \hline
\textcolor{OrangeRed}{bus}     & \textcolor{blue}{bench}                      & \textcolor{MidnightBlue}{kite}               &  \textcolor{Mahogany}{cell phone}\\ \hline
\textcolor{OrangeRed}{car}     & \textcolor{Fuchsia}{chair}                       &  \textcolor{MidnightBlue}{baseball bat}         &  \textcolor{BlueViolet}{book} \\ \hline
\textcolor{LimeGreen}{cat}     & \textcolor{Fuchsia}{diningtable}            & \textcolor{MidnightBlue}{baseball glove}             & \textcolor{BlueViolet}{clock} \\ \hline
\textcolor{LimeGreen}{cow}     & \textcolor{Fuchsia}{pottedplant}                    & \textcolor{MidnightBlue}{skateboard}      & \textcolor{BlueViolet}{vase} \\ \hline 
\textcolor{LimeGreen}{dog}     & \textcolor{YellowOrange}{backpack}             & \textcolor{MidnightBlue}{surfboard}           &  \textcolor{BlueViolet}{scissors}\\ \hline
\textcolor{LimeGreen}{horse}     & \textcolor{YellowOrange}{umbrella}            & \textcolor{MidnightBlue}{tennis racket}          &  \textcolor{BlueViolet}{teddy bear}\\ \hline
\textcolor{OrangeRed}{motorbike}     & \textcolor{YellowOrange}{handbag}            & \textcolor{Green}{banana}                 &  \textcolor{BlueViolet}{hair drier} \\ \hline
\textcolor{LimeGreen}{sheep}     & \textcolor{YellowOrange}{tie}                 & \textcolor{Green}{apple}                         &  \textcolor{BlueViolet}{tootbrush}  \\ \hline
\textcolor{OrangeRed}{train}     &   \textcolor{YellowOrange}{suitcase}                                     & \textcolor{Green}{sandwhich}                      & \textcolor{Thistle}{wine glass} \\ \hline
\textcolor{LimeGreen}{elephant}     & microwave                                 & \textcolor{Green}{orange}                   &  \textcolor{Thistle}{cup}\\ \hline
\textcolor{LimeGreen}{bear}     & oven                                          &  \textcolor{Green}{broccoli}                 &  \textcolor{Thistle}{fork} \\ \hline  
\textcolor{LimeGreen}{zebra}     & toaster                                       & \textcolor{Green}{carrot}                     &  \textcolor{Thistle}{knife} \\ \hline
\textcolor{LimeGreen}{giraffe}     & sink                                        & \textcolor{Green}{hot dog}                            &   \textcolor{Thistle}{spoon} \\ \hline
\textcolor{OrangeRed}{truck}     & refrigerator                                  & \textcolor{Green}{pizza}                  &  \textcolor{Thistle}{bowl} \\ \hline
\textcolor{Gray}{person}     & \textcolor{Fuchsia}{bed}                                 & \textcolor{Green}{donut}                      &  \textcolor{Mahogany}{tvmonitor} \\ \hline
                              & \textcolor{Fuchsia}{toilet}                                       &  \textcolor{Green}{cake}      & \textcolor{Thistle}{bottle} \\ \hline
                             & \textcolor{Fuchsia}{sofa  }                                            &        &  \\ \hline
\end{tabular}

\centering

\vspace{+5mm}
\caption{\textbf{Composition of OWDETR Split.} There is no overlap of categories between each task}
\label{tab:owdetr_split}
\end{table}

\clearpage 

\begin{figure*}[h!]
  \begin{center}
    \includegraphics[width=0.6\textwidth]{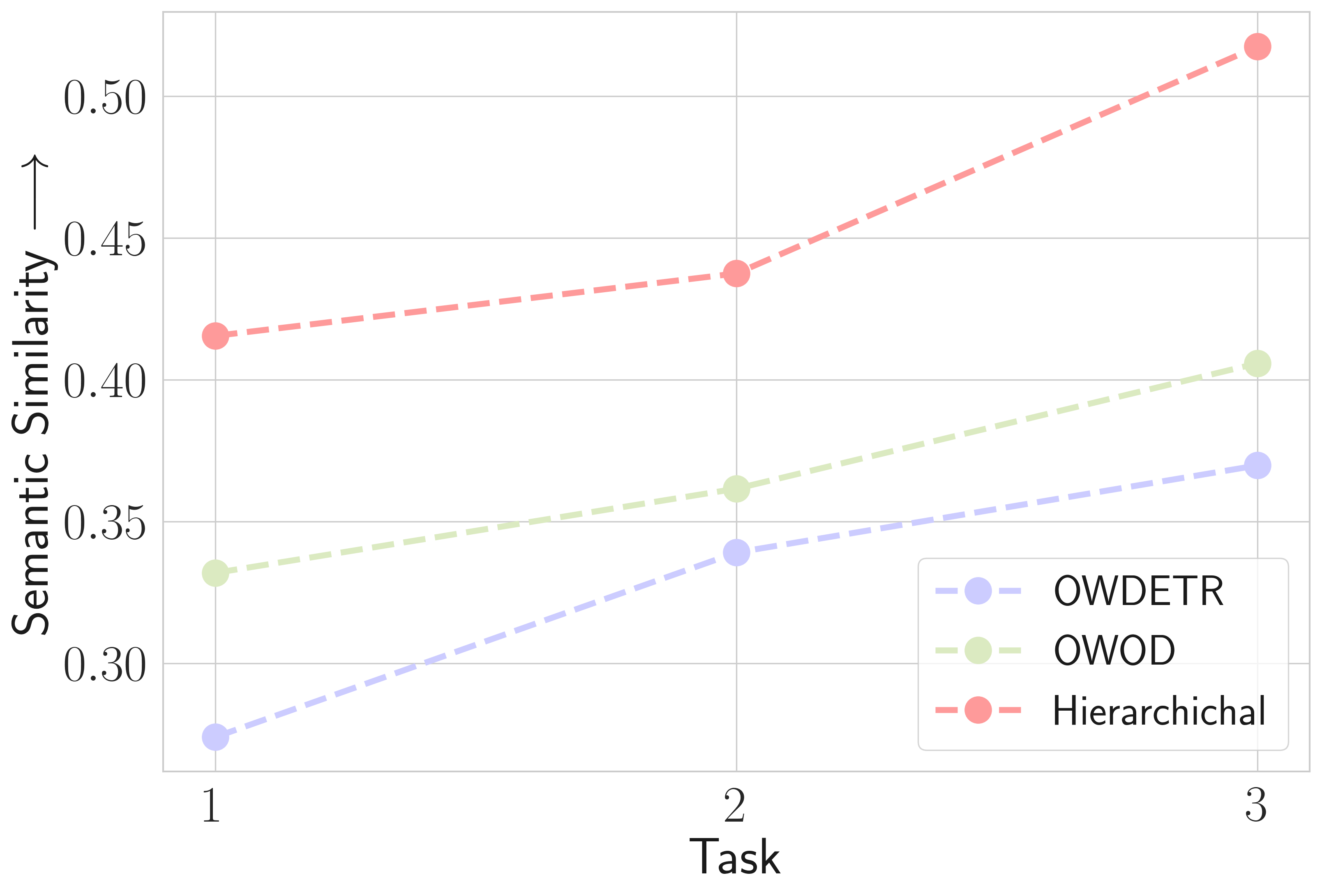}
    % \rule{3cm}{7cm}
  \end{center}
   % \vspace{+32mm}
  \caption{\textbf{Semantic Similarity between knowns and unknowns across tasks for each Split.}}
  \label{fig:dataset_similarity}
\end{figure*}

\paragraph{Quantifying Semantic Overlap between Knowns and Unknowns across Dataset} \label{app:similarity_measure}
We propose a measure to quantify the semantic similarity overlap between known and unknown classes in each dataset.  To achieve this, we utilize the GloVe \citep{pennington-etal-2014-glove} embedding for each known class $c \in \cK^{t}$ (and unknown class $k \in \cU^{t}$), denoted by $\mb{\omega_{c}}$ ( respectively $\mb{\omega_{k}}$ ). For a given task $t \in [1,T-1]$, we define the semantic overlap between knowns and unknowns as:
\begin{align}
S_{t}=\frac{1}{|\cU^{t}|} \displaystyle{\sum_{k \in \cU^{t}}  \max_{c \in \cK^{t}}} 
\frac{ < \mb{\omega^{c}},\mb{\omega^{k} >}}  {\lVert \mb{\omega^{c}}  \rVert \cdot \lVert \mb{\omega^{k}}  \rVert}  , \forall t \leq T-1
\label{eq:semantic_measure}
\end{align}
Higher value indicates larger similarity between the knowns and unknowns. Figure~\ref{fig:dataset_similarity} shows the evolution of this similarity measure throughout the training. This measure quantifies the similarity overlap between known and unknown items, with higher values indicating larger overlap. The three splits, OW-DETR Split (\textcolor{Periwinkle}{Low regime}), OWOD Split (\textcolor{LimeGreen}{Medium regime}), and Hierarchical Split (\textcolor{Red}{High regime}), consistently align with our intended design, providing a foundation for evaluating baseline methods across diverse scenarios. Note that this metric is increasing as the number of knowns grows throughout the training.

\newpage

\section{Implementation Details} \label{app:implementation_details}
In this section, we provide a recap of the definition and value of each hyperparameter used, as well as the learning rate schedule employed during training. The sequence of learning rate for each task is recalled in Table~\ref{tab:learning_rate_schedule}. Additional details are provided regarding the usage of the buffer in various loss functions.

\begin{table}[h!]
\centering

\begin{tabular}{c|c|c}
\hline
\multicolumn{3}{c}{ \textbf{Hyperparameters}} \\ \hline
Parameters  & Value & Definition   \\ \hline
   $\alpha$  & 0.05   & Coefficient of the Hyperbolic Contrastive Loss Eq~\ref{eq:contrastive_loss}   \\ \hline
   $\beta$ &   0.02   & Coefficient of the SuperClass Regularizer Eq~\ref{eq:family_regularizer}  \\  \hline
    $\tau_1$ &   0.2    & Temperature of the   Hyperbolic Contrastive Loss \\  \hline
    $\tau_2$ &      0.4  & Temperature of the   SuperClass Regularizer\\  \hline
    $m$     &      10  & Capacity of the replay buffer (examplar per class)  \\  \hline
      $c$    &     0.1   &  curvature coefficient for the Hyperbolic Distance  \\  \hline
       $k$    &     1  &  number of positive examples to be sampled for each anchor \\  \hline
        $P$    &     12  &  There are a total of 12 categories defined originally by \citet{mscoco} \\  \hline
      % $l_r$    &     $10^{-4}$    & then decreases to $10^{-5}$ \\  \hline
      batch size &   3     & N/A\\  \hline
       GPUs &     4 Nvidia RTX 3090     & N/A \\  \hline
\end{tabular}

\centering

\vspace{+5mm}
\caption{\textbf{Hyperparameters used for \method.}}
\label{tab:hyperparameters}
\end{table}

\begin{table}[h!]
\centering

\begin{tabular}{c|c|c|c}
\hline
\multicolumn{4}{c}{ \textbf{Hierarchical Split}} \\ \hline
Task  $\cT_{t}$  & From Epoch to & Learning rate $l_r$  & Classes labeled \\ \hline
   Task 1  & 0 to 40 &    $10^{-4}$ & 0-19 \\
   Task 1  & 40 to 50 &    $10^{-5}$ &  0-19 \\   \hline
   Task 2 &   50 to70     & $10^{-4}$&  20-39\\  
    Task 2 fine-tuning &   70 to 130     & $10^{-4}$  &  0-39 \\  \hline
    Task 3 &      130 to150  &  $10^{-4}$ &   40-59 \\ 
    Task 3 fine-tuning   & 150 to210      &  $10^{-4}$ & 0-59  \\  \hline
      Task 4     &     210 to 230    &  $10^{-4}$ & 60-79 \\  
      Task 4 fine-tuning    &     230-300    &  $10^{-4}$ &  0-80 (no unknowns)\\  \hline
     
\end{tabular}

\centering
\vspace{+5mm}
\caption{\textbf{Learning rate schedule throughout the 4 tasks.}}
\label{tab:learning_rate_schedule}
\end{table}

\begin{table}[h!]
\centering

\begin{tabular}{c|c|c}
\hline
Parameters  & Range & Comments \\ \hline
$\alpha$   & $\{ 0.05, 0.1, 0.2 \}$  & \\
     $\beta$   & $\{ 0.05, 0.1, 0.2 \}$  & should be lower than $\alpha$ \\
     $\tau_2$   & $\{ 0.02, 0.05, 0.1 \}$  &  should be lower than $\tau_1$ as in~\citep{li2023updp}\\
   $m$   & $\{ 10,20,50 \}$  & Negligible impact, yet increases memory \\
    $c$   & $\{ 0.0,0.1,0.2,0.5 \}$ & See Table~\ref{tab:ablation_curvature} \\
    $k$   & $\{ 1,2,5 \}$  & Negligible impact, yet increases training time  \\ \hline
\end{tabular}

\centering
\vspace{+5mm}
\caption{\textbf{Hyperparameters Tuning Ranges.}}
\end{table}

\paragraph{Buffer Filling:}
Throughout the training process, we store the embeddings of every encountered class $\mb{z_{i}^{c}}$ (where $c \in \cK, i \in \mathcal{B}$) which has a capacity of $m$ exemplars per class.

\paragraph{Buffer Interaction with Hyperbolic Contrastive Loss:}
To compute the contrastive loss (Eq.~\ref{eq:contrastive_loss}), for each embedding $z_{i}^{c}$ in the batch, where $i \in \cB$ and $c \in \cK$ , we sample its positive counterpart $\mb{z_{i^{+}}}$ from the same class $c$ in the buffer.

\paragraph{Buffer Interaction with SuperClass Regularizer:}
To compute the Hyperbolic Average $\mb{\overline{z}_{p}}$ ($p=1...P$) in the SuperClass Regularizer (Eq.~\ref{eq:family_regularizer}), we sample from the buffer every embedding $\mb{z_{i}^{c}}$, $i \in M, c \in \cS_{p}$ then use Eq~\ref{eq:category_embedding} to calculate $\mb{\overline{z}_{p}}$.

\newpage

\section{Additional Experimental Results} \label{app:detailed_results}

\subsection{Detailed Results}
Table~\ref{tab:final_results} presents a detailed overview of the performance of all baselines across the three splits. It includes the mean Average Precision (mAP) of the Previous and Current known classes for tasks 2, 3, and 4. \method exhibits a high level of plasticity, as evidenced by its accuracy on the current known classes (highlighted in the 'Current' column of tasks 2, 3, and 4). This can be attributed to the hierarchical structure learned within the hyperbolic embedding space, which facilitates effective modeling and representation of the known classes.
\begin{table*}[ht!]
\resizebox{1.0\textwidth}{!}{
\begin{tabular}
{c|l|cc||cccc||cccc||ccc}\hline
\parbox[t]{3mm}{\multirow{3}{*}{\rotatebox[origin=c]{90}{\textbf{Regime}}}} &  & \multicolumn{2}{c}{\textbf{Task 1}} & \multicolumn{4}{c}{\textbf{Task 2}}& \multicolumn{4}{c}{\textbf{Task 3}} & \multicolumn{3}{c}{\textbf{Task 4}} \\  \cline{3-15} 
 &     &  U-Recall & mAP ($\uparrow$)  &  U-Recall ($\uparrow$) &  \multicolumn{3}{c||}{mAP ($\uparrow$)}  & U-Recall  ($\uparrow$) &  \multicolumn{3}{c||}{mAP ($\uparrow$)} & \multicolumn{3}{c}{mAP ($\uparrow$)} \\ 
  &   Methods            &     &             Current         &         &    Previous & Current & Both                  &   & Previous & Current & Both  & Previous & Current & Both                                    \\ \hline
\parbox[t]{3mm}{\multirow{4}{*}{\rotatebox[origin=c]{90}{\textcolor{Periwinkle}{\textbf{Low}}}}}&  ORE - EBUI           & 1.5  &  61.4 &   3.9 &   56.5 &   26.1  &  40.6 &   3.6  &  38.7  &  23.7  &  33.7  &  33.6 &   26.3 &   31.8 \\
&  OW-DETR              & 5.7 &71.5 & 6.2 & 62.8 & 27.5 & 43.8 & 6.9 & 45.2 &  24.9 & 38.5 & 38.2 & 28.1 & 33.1 \\
& PROB                               &    17.6 & \textbf{73.4} & 22.3 & 66.3 & 36.0 & 50.4 & 24.8 & 47.8 & 30.4 & 42.0 & 42.6 & 31.7 &  39.9 \\
 & \method (Ours)                                   & \textbf{23.9}   &   72.7   & \textbf{23.3}  &  59.8 & 42.2  & \textbf{50.6}   & \textbf{25.4}   & 49.3  & 39.8  & \textbf{46.2}   & 46.4 & 40.3   & \textbf{44.8}    \\   \hline
  &  $\Delta$(Rel. Difference)                            & \textcolor{Green}{\textbf{+6.3}} &  $\leq 1.0$   &  \textcolor{Green}{\textbf{+1.0}}  &   &   &   $\leq 1.0$ &  $\leq 1.0$  &  &  & \textcolor{Green}{$\boldsymbol{+4.2}$} &   &   &   \textcolor{Green}{$\boldsymbol{+4.9}$}     \\  \hline
\multirow{8}{*}{\rotatebox[origin=c]{90}{\textcolor{LimeGreen}{\textbf{Medium}}}} & ORE - EBUI    & 4.9   & 56.0   &  2.9  &  52.7  &  26.0  &  39.4  &   3.9 & 38.2 & 12.7 & 29.7 & 29.6 & 12.4 & 25.3   \\ 
&  UC-OWOD                    & 2.4   & 50.7   &  3.4  &  33.1  &  30.5  &  31.8  &  8.7  & 28.8 & 16.3 & 24.6 & 25.6 & 15.9 & 23.2   \\ 
&   OCPL                       & 8.26  & 56.6   &  7.65 &  50.6  &  27.5  &  39.1  &  11.9 & 38.7 & 14.7 & 30.7 & 30.7 & 14.4 & 26.7   \\ 
 & 2B-OCD                     & 12.1  & 56.4   &  9.4  &  51.6  &  25.3  &  38.5  &  11.6 & 37.2 & 13.2 & 29.2 & 30.0 & 13.3 & 25.8   \\ 
  &OW-DETR     & 7.5   & 59.2   &  6.2  &  53.6  &  33.5  &  42.9  &   5.7 & 38.3 & 15.8 & 30.8 & 31.4 & 17.1 & 27.8   \\ 
 &  PROB           & 19.4  & 59.5   & 17.4  &  55.7  &  32.2  &  44.0  &  19.6 & 43.0 & 22.2 & 36.0 & 35.7 & 18.9 & 31.5      \\
 &  \method (Ours)                             & \textbf{23.5}   & 59.4    & \textbf{20.6}   & 51.4  & 37.4   & \textbf{44.4}  &  \textbf{26.3}   & 42.9 & 24.6 & \textbf{36.8} &  37.4 &  22.4 &  \textbf{33.6}     \\   \hline
 &  $\Delta$(Rel. Difference)                            & \textcolor{Green}{$\boldsymbol{+4.1}$}  & $\leq 1.0$    &  \textcolor{Green}{$\boldsymbol{+3.1}$}  &   &   & $\leq 1.0$   &  \textcolor{Green}{$\boldsymbol{+6.6}$}  &  &  & $\leq 1.0$ &   &   &   \textcolor{Green}{\textbf{+2.1}}     \\   \hline
\parbox[t]{3mm}{\multirow{3}{*}{\rotatebox[origin=c]{90}{\textcolor{Red}{\textbf{High}}}}} & OW-DETR              & 7.0 &  47.3  & 11.0   &  38.0  & 39.2  &  38.6  &  8.8 &  39.3 & 36.1 & 38.3 & 38.5 & 37.3 & 38.2 \\
& PROB                                  & 29.4 & 49.6  & 43.9  & 41.9   &  43.9  & 42.9   & 52.7  & 41.7 & 40.4 & 41.3 & 40.9  & 41.2 &  41.0\\
 & \method (Ours)                 &  \textbf{34.9}  & \textbf{49.9}    & \textbf{47.5}  & 42.0    & 49.0  & \textbf{45.5}   &  \textbf{55.2}   & 44.4 & 44.1 & \textbf{44.3}  & 42.8 &  47.0 &  \textbf{43.9} \\  \hline
 &  $\Delta$(Rel. Difference)                            & \textcolor{Green}{\textbf{+5.5}} &  $\leq 1.0$   &  \textcolor{Green}{$\boldsymbol{+3.6}$}  &   &   &   \textcolor{Green}{$\boldsymbol{+2.6}$}  &  \textcolor{Green}{$\boldsymbol{+2.5}$}  &  &  & \textcolor{Green}{$\boldsymbol{+3.0}$} &   &   &   \textcolor{Green}{$\boldsymbol{+2.9}$}     \\  \hline
\end{tabular}
}
\caption{\textbf{Benchmark Results.}}
\label{tab:final_results}
\end{table*}

Next, we turn our attention to the issue of unknown object confusion and address it through the Absolute Open-Set Error (A-OSE) metric, as introduced by \cite{owod}. The A-OSE quantifies the number of unknown objects that are incorrectly classified as known (lower is better). In Table~\ref{tab:unknown_confusion}, we present the A-OSE values alongside the U-Recall metric for the Hierarchical Split.

\begin{table*}[ht!]
\centering
% \hspace{-14mm}
\resizebox{1.0\textwidth}{!}{
\begin{tabular}
{c|l|cc||cc||cc}  \hline
  \multirow{2}{*}{\rotatebox[origin=c]{90}{\tiny{\textbf{Regime}}}}  &  &\multicolumn{2}{c}{\textbf{Task 1}}  &\multicolumn{2}{c}{\textbf{Task 2}} &\multicolumn{2}{c}{\textbf{Task 3}} \\    \cline{3-8}
    % \hline
    &  Methods  &  U-Recall ($\uparrow$)  & A-OSE($\downarrow$)  &  U-Recall ($\uparrow$)  & A-OSE($\downarrow$) &  U-Recall ($\uparrow$)  & A-OSE($\downarrow$)  \\ \hline
\multirow{3}{*}{\rotatebox[origin=c]{90}{\small\textcolor{LimeGreen}{\textbf{Medium}}}} &   OW-DETR                 &  7.5  &  10,240     &  6.2 &  8,441 & 5.7 & 6,803\\   
 &  PROB              &  19.4 & \textbf{5,195}    &  17.4 &  6,452  &  19.6&    \textbf{2,641} \\ 
 &   \method (Ours)             &  \textbf{23.5}   &  11,275    &  \textbf{20.6} & \textbf{4,805} & \textbf{26.3} & 3,548  \\ \hline 
  \parbox[t]{3mm}{\multirow{3}{*}{\rotatebox[origin=c]{90}{\small\textcolor{Red}{\textbf{High}}}}} &   OW-DETR                 &  7.0  & 42,540     &  11.0 & 26,527 & 8.8 & 20,034\\   
 &  PROB              &  29.4 & 14,962    &  43.9 & 8,929 &  52.7&   5,387 \\ 
  &  \method (Ours)             &  \textbf{34.9}   &  \textbf{7,420}    &  \textbf{47.5} & \textbf{3,849} & \textbf{55.2} & \textbf{4,611}  \\  \hline
\end{tabular}
}
\caption{\textbf{Unknowns Confusion.} \method achieves the highest U-Recall, indicating its superior ability to detect unknown objects. The misclassification rate (A-OSE) shows improved performance for \method in the high regime, where inherent hierarchical structure is more pronounced than the medium regime.}
\label{tab:unknown_confusion}
\end{table*}

\clearpage
\subsection{Qualitative Results}
\begin{figure*}[ht!]
  \begin{center}
    \includegraphics[width=1.0\textwidth]{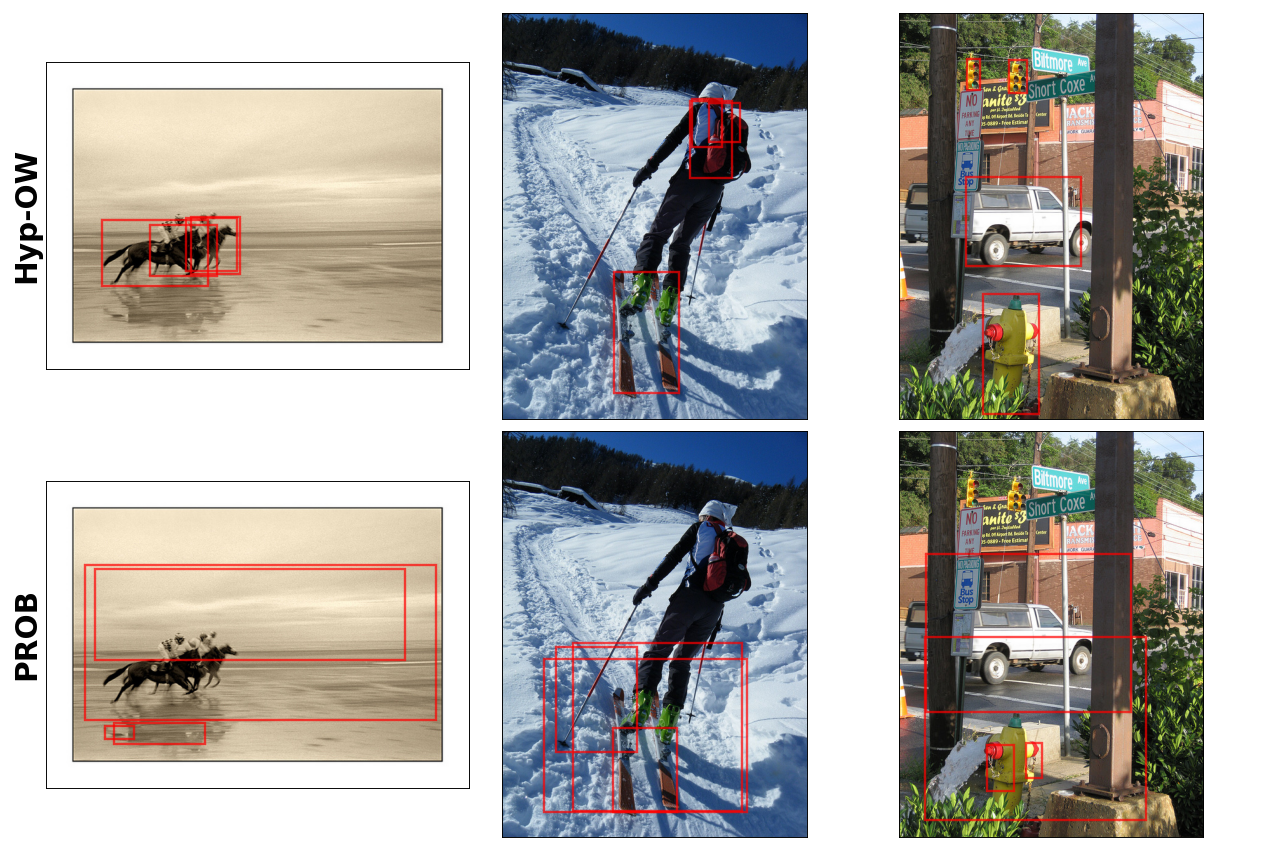}
    % \rule{3cm}{7cm}
  \end{center}
   % \vspace{+32mm}
  \caption{\textbf{Unknown detection illustrated for Hyp-OW (first row) and PROB (second row), highlighted with the \textcolor{red}{red boxes}.} By learning the hierarchical structure of items, Hyp-OW provides a more fine-grained detection of objects. On the other hand, PROB, which learns a generalized objectness score by averaging all features, is restricted to a more broad, coarse-grained detection approach}
  \label{fig:HYPOW_versus_PROB}
\end{figure*}

\begin{figure*}[ht!]
  \begin{center}
    \includegraphics[width=1.0\textwidth]{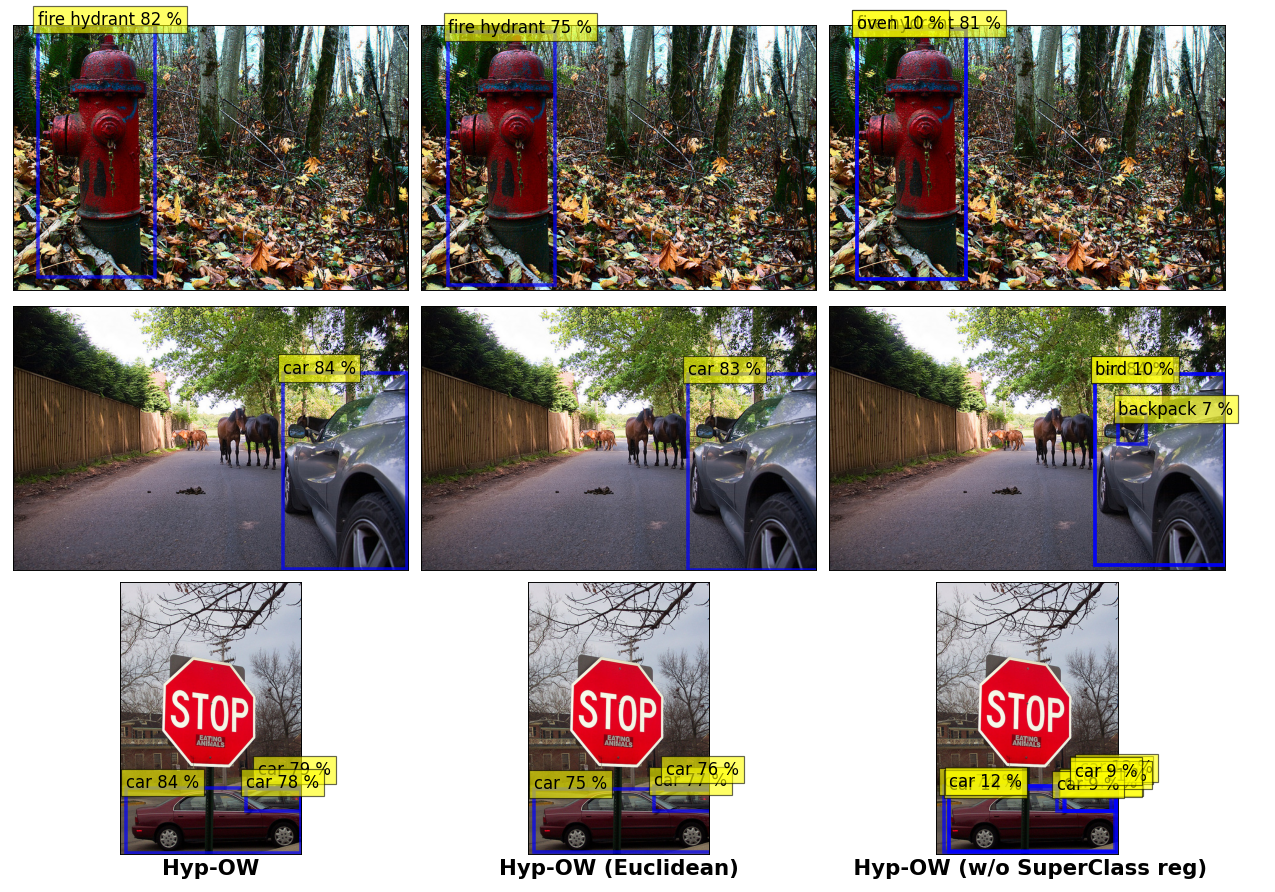}
    % \rule{3cm}{7cm}
  \end{center}
   % \vspace{+32mm}
  \caption{\textbf{Impact of each component of Hyp-OW on known items accuracy.} Hyp-OW (left column) exhibits not only higher confidence but also superior capability in identifying correct items.}
  \label{fig:HYPOW_ablation}
\end{figure*}

In this section, we present a qualitative visualization that clearly illustrates the superior detection precision of our method, Hyp-OW, in comparison to the PROB baselines (Figure~\ref{fig:HYPOW_versus_PROB}). Additionally, we provide a detailed quantitative assessment underscoring the impact of each individual component within the Hyp-OW framework (Figure~\ref{fig:HYPOW_ablation}).

\clearpage
\subsection{Unveiling Co-occurrence Learning: The Influence of Bounding Box Overlap}
\begin{figure*}[ht!]
  \begin{center}
    \includegraphics[width=1.0\textwidth]{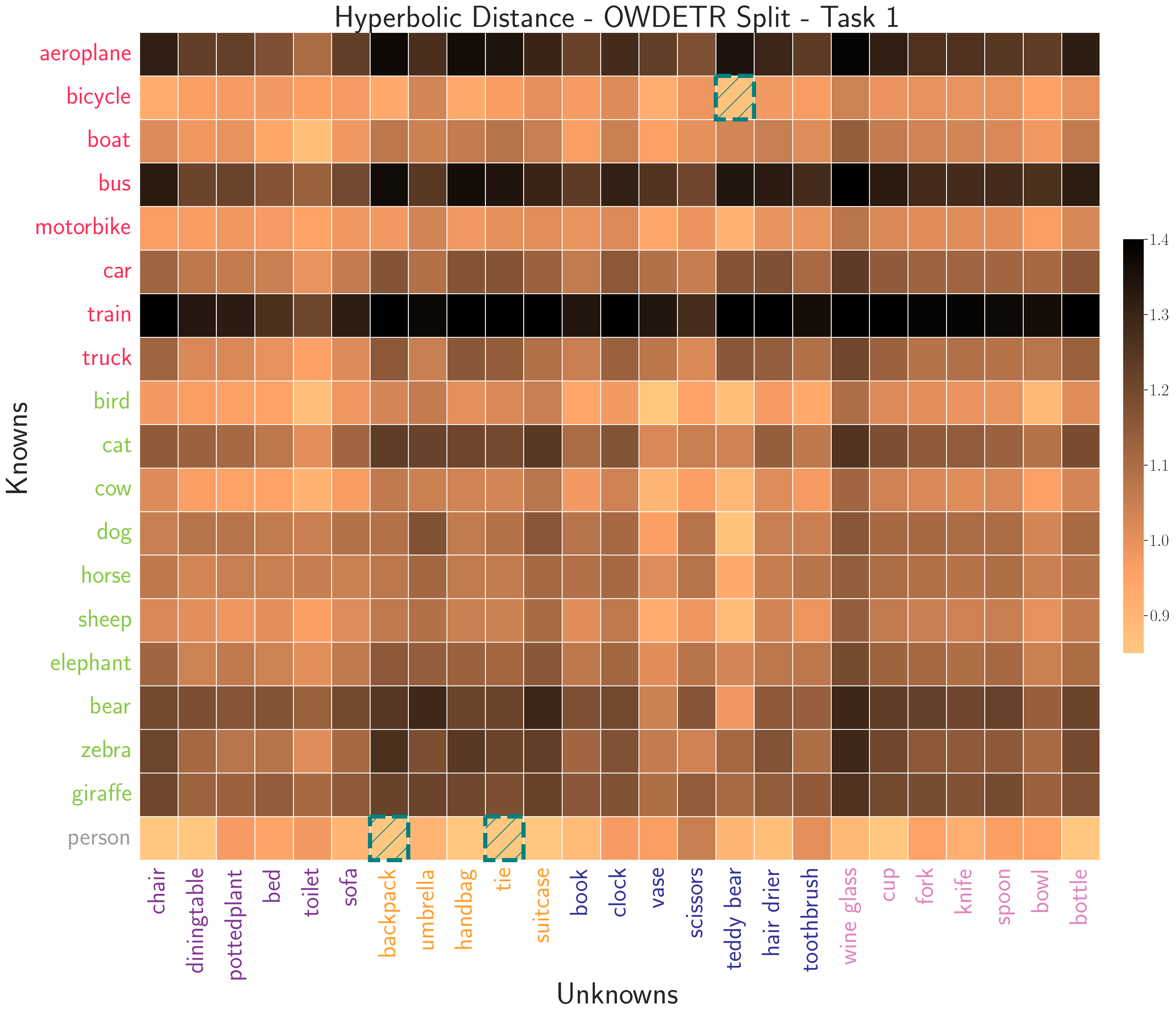}
  \end{center}
  \caption{\textbf{Hyperbolic Knowns-Unknowns Distance Heatmap.} Lighter colors indicate lower hyperbolic distance and higher similarity. In the presence of images with high bounding box overlap, the model can learn frequent associations, such as 'person' and 'tie' or 'backpack' (teal hatched boxes, last row) or 'teddy bear' and 'bicycle'.}
  \label{fig:heatmap_counter_example_owdetr}
\end{figure*}

\begin{figure*}
  \begin{center}
    \includegraphics[width=1.0\textwidth]{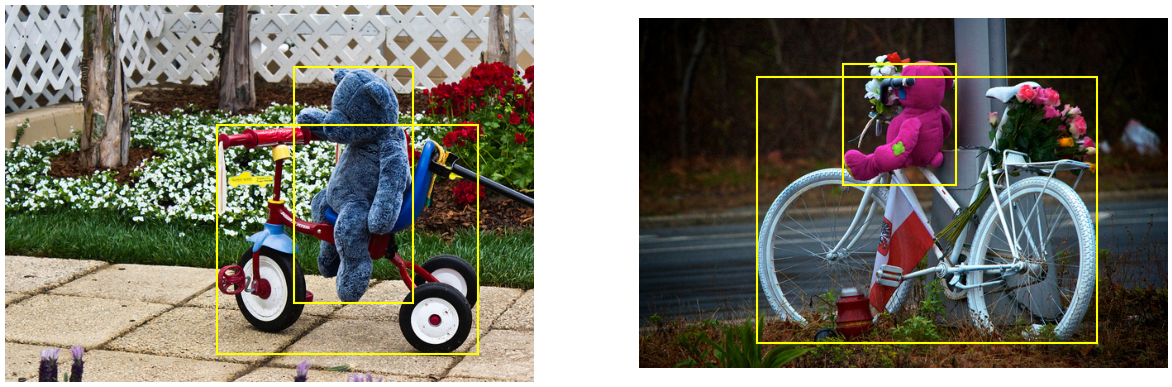}
  \end{center}
  \caption{\textbf{Bounding boxes overlap between classes 'teddy bear' and 'bicycle'.}}
  \label{fig:bbox_overlap_teddybear_bicycle}
\end{figure*}

\begin{figure*}[ht!]
  \begin{center}
    \includegraphics[width=1.0\textwidth]{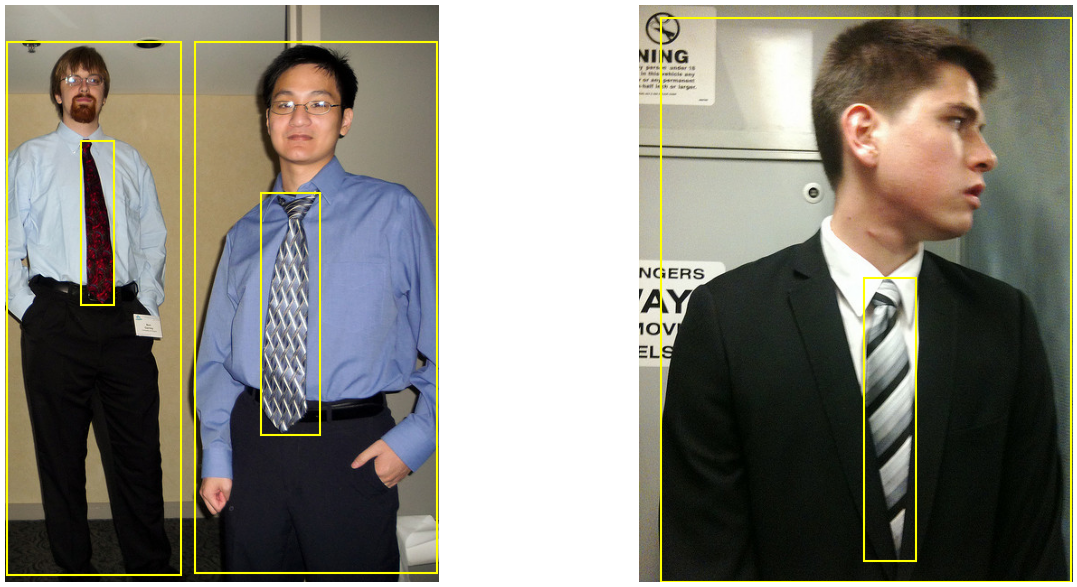}
    % \rule{3cm}{7cm}
  \end{center}
   % \vspace{+32mm}
  \caption{\textbf{Bounding boxes overlap between classes 'person' and 'tie'.}}
  \label{fig:bbox_overlap_person_tie}
\end{figure*}

\begin{figure*}[h!]
  \begin{center}
    \includegraphics[width=1.0\textwidth]{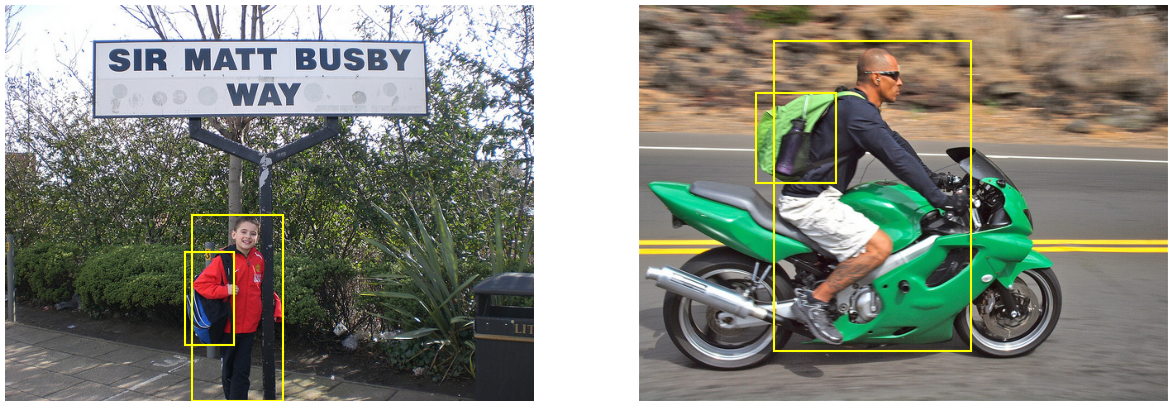}
    % \rule{3cm}{7cm}
  \end{center}
   % \vspace{+32mm}
  \caption{\textbf{Bounding boxes overlap between classes 'person' and 'backpack'.}}
  \label{fig:bbox_overlap_person_backpack}
\end{figure*}

In the Object Detection (OD) task, where bounding boxes can overlap, the model has a tendency to learn co-occurrence patterns of frequently appearing classes. This phenomenon is quantitatively demonstrated in the Heatmap Figure~\ref{fig:heatmap_counter_example_owdetr}, where the model exhibits high similarity between classes that do not have any semantic relationship, such as 'person' and 'tie', or 'teddy bear' and 'bicycle' (highlighted in teal). This observation is further supported by Figures  ~\ref{fig:bbox_overlap_teddybear_bicycle}, ~\ref{fig:bbox_overlap_person_tie} and ~\ref{fig:bbox_overlap_person_backpack}, which provide qualitative examples of this co-occurrence learning.

\newpage

\clearpage 
\section{Ablation Analysis}  \label{app:ablation}
In this section, we present comprehensive results of the ablation analysis. We then delve into a detailed discussion of the impact of each component.

\subsection{Detailed Ablation Results}
To analyze the impact of each component, we systematically deactivate individual modules and observe the resulting performance (Table~\ref{tab:detailed_appendix}). In summary, deactivating the Adaptive Relabeling module leads to a significant decrease in known class detection (mAP) while maintaining a stable U-Recall. On the other hand, the SuperClass Regularizer and Hyperbolic Distance modules are responsible for the U-Recall performance. These modules play a crucial role in achieving a balance between known class detection (mAP) and unknown class retrieval (U-Recall) maintaining an equilibrium in the detection process.

\begin{table*}[ht!]
% \centering
% \hspace{-14mm}
\resizebox{1.0\textwidth}{!}{
\begin{tabular}
{c|l|cc||cccc||cccc||ccc} \hline
 \parbox[t]{3mm}{\multirow{3}{*}{\rotatebox[origin=c]{90}{\textbf{Regime}}}} &  & \multicolumn{2}{c}{\textbf{Task 1}} & \multicolumn{4}{c}{\textbf{Task 2}}& \multicolumn{4}{c}{\textbf{Task 3}} & \multicolumn{3}{c}{\textbf{Task 4}} \\ \cline{3-15} 
  &     &  U-Recall ($\uparrow$)  & mAP($\uparrow$)   &  U-Recall ($\uparrow$)  &  \multicolumn{3}{c||}{mAP ($\uparrow$) }  & U-Recall ($\uparrow$)  &  \multicolumn{3}{c||}{mAP ($\uparrow$) } & \multicolumn{3}{c}{mAP ($\uparrow$) } \\ 
   &                  &       &             Current         &              &    Previous & Current & Both                  &  & Previous & Current & Both  & Previous & Current & Both                                    \\ \hline
\parbox[t]{3mm}{\multirow{4}{*}{\rotatebox[origin=c]{90}{\textcolor{LimeGreen}{\textbf{Medium}}}}} & \method (Ours)                 & \textbf{23.5}   & \textbf{59.4}    & \textbf{20.6}  & 51.4  & 37.4   & \textbf{44.4}  &  26.3   & 42.9 & 24.6 & \textbf{36.8} &  37.4 &  22.4 &  33.6     \\ 
& with {\fontfamily{qcr}\selectfont Cosine Distance} ($c=0$)            & 21.7  &  59.1 & 19.2  & 51.6  & 33.6 & 42.6  & 25.5   & 40.5 & 23.9 & 35.0  & 36.7 & 20.7 & 32.7  \\
& w/o {\fontfamily{qcr}\selectfont SuperClass Regularizer} ($\beta=0$)            & 22.6 & 59.0 & 19.0 & 50.4 &36.8 & 43.6  & 23.7  & 41.4 &  24.7 & 35.8  & 37.8 & 22.0 & \textbf{33.8} \\
& w/o  {\fontfamily{qcr}\selectfont Adaptive Relabeling}          & 22.7 & 44.6  &  18.7  & 38.5 &   28.1& 33.3  &  \textbf{27.2} &  32.7& 40.7 &  28.8 & 29.8 &  19.3& 32.1 \\ \hline
\parbox[t]{3mm}{\multirow{4}{*}{\rotatebox[origin=c]{90}{\textcolor{Red}{\textbf{High}}}}} & \method (Ours)                 &  \textbf{34.9}  & 49.9    & 47.5  & 42.0    & 49.0   & \textbf{45.5}   &  55.2   & 44.4 & 44.1 & \textbf{44.3}  & 42.8 &  47.0 &  \textbf{43.9}  \\ 
& w/ {\fontfamily{qcr}\selectfont Cosine Distance} ($c=0$)          & 32.8  & 49.0  &  46.4   & 42.0 & 48.8 & 45.4 & \textbf{55.4} & 43.6  & 42.3 &  43.2  & 42.4 & 45.2 & 43.1\\
& w/o {\fontfamily{qcr}\selectfont SuperClass Regularizer} ($\beta=0$)        & 32.0   & \textbf{50.0} &  47.1 & 41.3 &  48.9 & 45.1 & 52.9   & 43.8 & 43.5 & 43.7 & 42.8 & 45.9 & 43.5 \\
& w/o  {\fontfamily{qcr}\selectfont Adaptive Relabeling}          & 34.7 & 41.2   & \textbf{47.6}  & 36.2   & 41.5   & 38.9      & 54.1  & 36.5  & 36.6  & 36.5   &34.5 & 37.9 & 36.1    \\ \hline 
\end{tabular}
}
\caption{\textbf{Impact of each component of \method on OWOD and Hierachichal Split.}}
\label{tab:detailed_appendix}
\end{table*}

\paragraph{Deactivating Relabeling Module:} To asses the impact of this module, we adopt PROB's~\citep{prob} methodology where every unmatched query $\mb{q^{\textit{u}}}$ is labelled as unknowns.

\paragraph{Deactivating SuperClass Regularizer:} This is done by setting $\beta=0$

\paragraph{Deactivating Hyperbolic Embedding:} This is done by setting $c=0$ 
\\
We here demonstrate how Euclidian distance and cosine similarity are linked in case of normalized vectors.
Recall that the Euclidian distance is recovered when $c \to 0$: $\lim_{c \to 0} d_{hyp}(\mb{x},\mb{y}) = 2 \rVert \mb{x} - \mb{y} \rVert$

For two normalized vectors $\mb{x},\mb{y}$:
\begin{align}
2^{2}\lVert \frac{\mb{x}}{\mb{\lVert x \lVert}} - \frac{\mb{y}}{\mb{ \lVert y \lVert}} \lVert^{2} = 4 (\lVert \frac{\mb{x}}{\mb{\lVert x \lVert}} \lVert^{2}  + \lVert  \frac{\mb{y}}{\mb{\lVert y \lVert}} \lVert 
^{2}-2\frac{ < \mb{x},\mb{y} >}{\lVert \mb{x}  \rVert \cdot \lVert \mb{y}  \rVert} )=4 (2-2\frac{ < \mb{x},\mb{y} >}{\lVert \mb{x}  \rVert \cdot \lVert \mb{y}  \rVert})= 4 \times d_{cos}(\mb{x},\mb{y}) 
\label{eq:cosine_euclidian}
\end{align}

\clearpage
\subsection{Effectiveness of our Semantic Adaptive Relabeling Scheme}
\begin{figure*}[ht!]
  \begin{center}
    \includegraphics[width=0.95\textwidth]{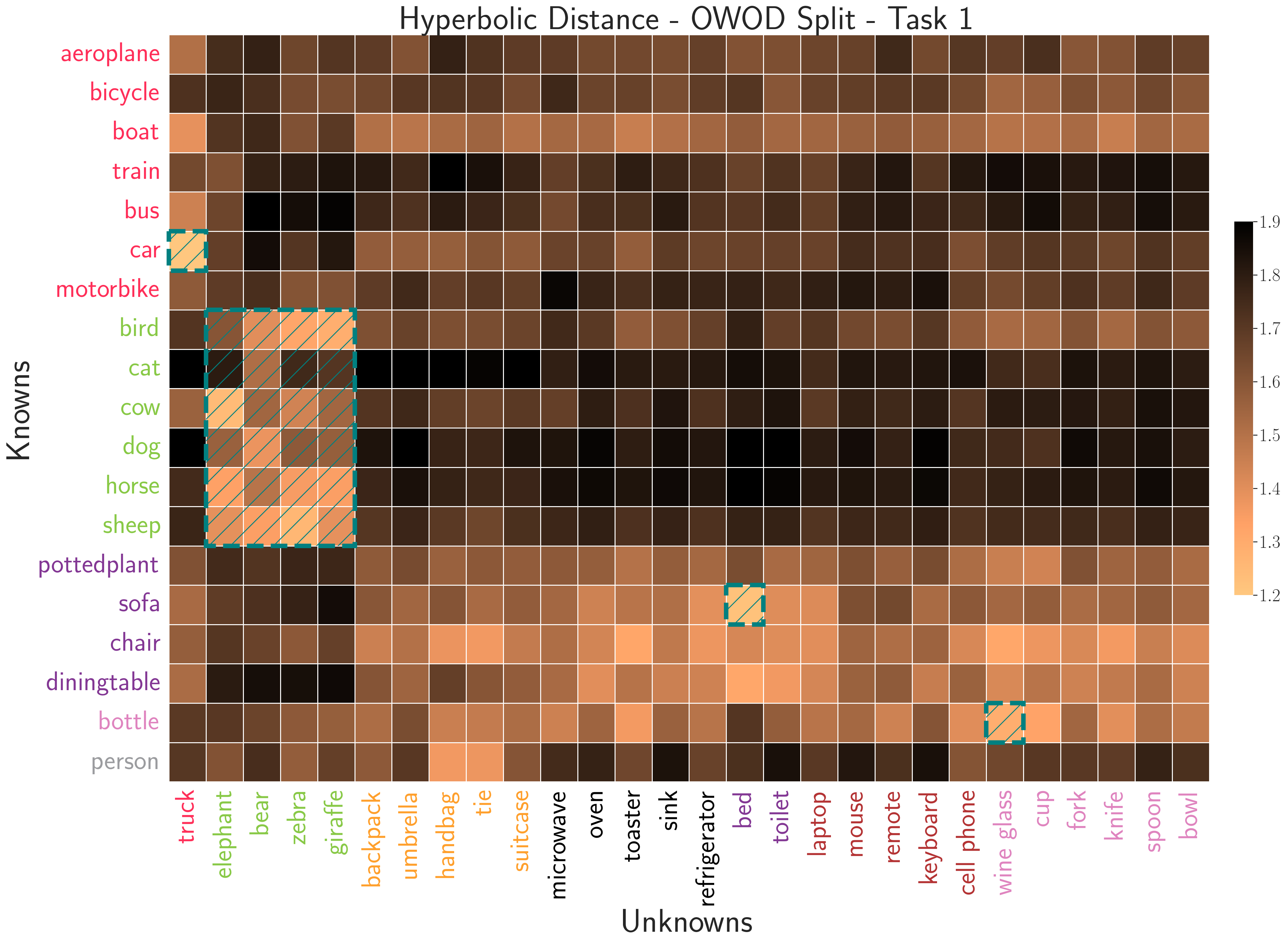}
  \end{center}
  \caption{\textbf{Hyperbolic Knowns-Unknowns Distance Heatmap}: In the heatmap, lighter colors indicate higher similarity. We observe that classes belonging to the same category, such as 'car' and 'truck' or 'bed' and 'sofa', exhibit higher similarity.}
  \label{fig:owod_semantic_similarity}
\end{figure*}

\begin{figure*}[ht!]
  \begin{center}
    \includegraphics[width=1.0\textwidth]{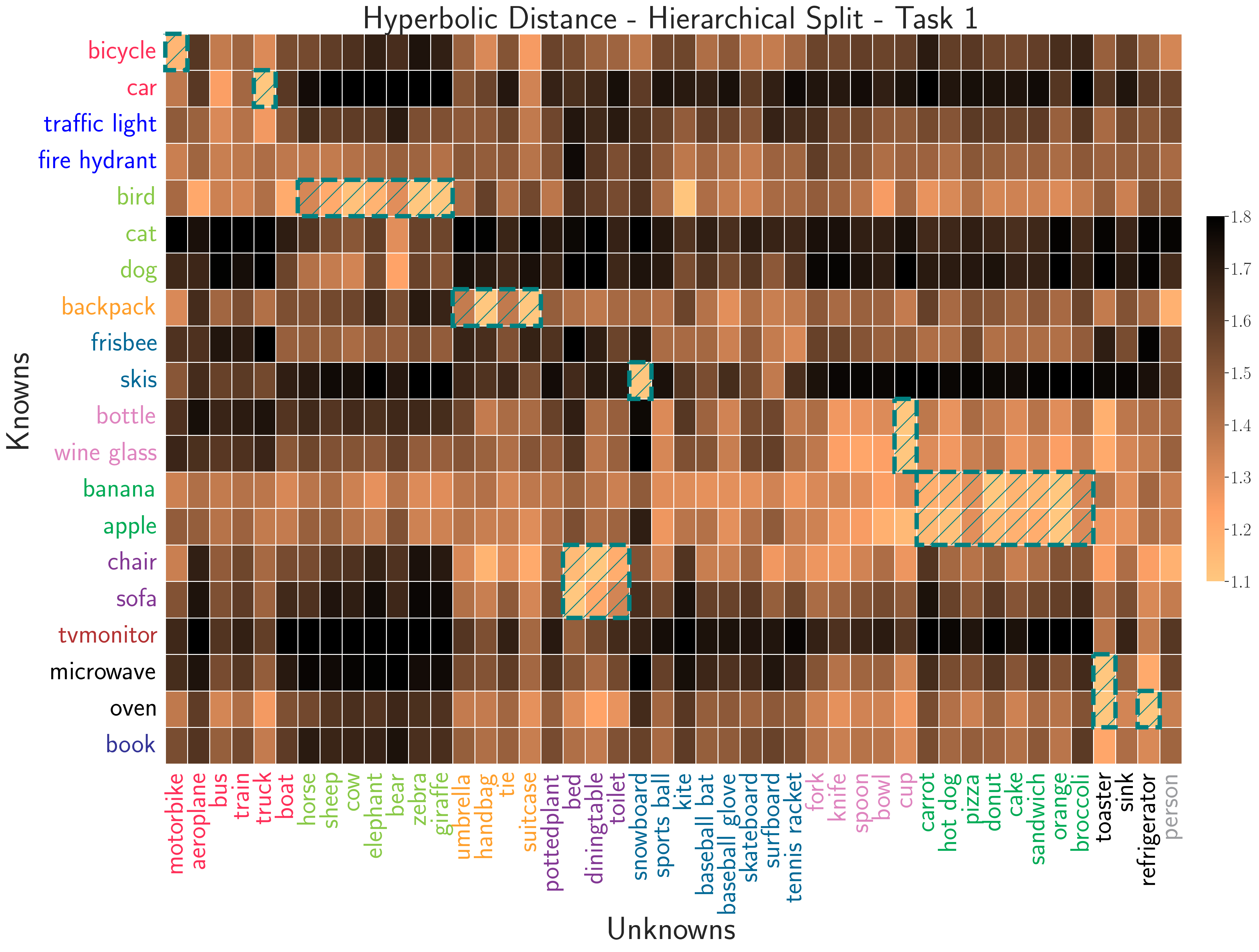}
  \end{center}
  \caption{
\textbf{Hyperbolic Knowns-Unknowns Distance Heatmap}: Lighter colors indicate higher similarity. We can observe that classes from the same category, such as 'skis' and 'snowboard' or 'toaster' and 'microwave', exhibit higher similarity.}
  \label{fig:hierarhichal_semantic_similarity}
\end{figure*}

We now provide insights into the effectiveness of our Semantic Distance-based Relabeling Scheme. Figure~\ref{fig:owod_semantic_similarity} and ~\ref{fig:hierarhichal_semantic_similarity} show the heatmap distance between known (left) and unknown items (bottom) for Task 1 in the OWOD and Hierarchical Split, respectively. Lighter colors indicate lower distances and higher similarity between known and unknown classes.

We can observe a significant similarity between known and unknown classes belonging to the same category, as highlighted by teal hatched boxes in both heatmaps. For example, in the OWOD Split, the animal classes from the knowns exhibit high similarity with the unknown classes. This is observed by comparing the classes on the left side of the heatmap, such as bird, cat, and cow, with the classes at the bottom, such as elephant and bear. In the Hierarchical Split, classes from the food category also demonstrate high similarity (see banana, apple versus cake and orange).

These findings emphasize the effectiveness of our Hyperbolic Distance-based Relabeling Scheme in capturing and leveraging the hierarchical and semantic relationships between known and unknown classes.

\clearpage
\subsection{SuperClass Regularizer}
\begin{figure*}[ht!]
    \includegraphics[width=0.95\linewidth]{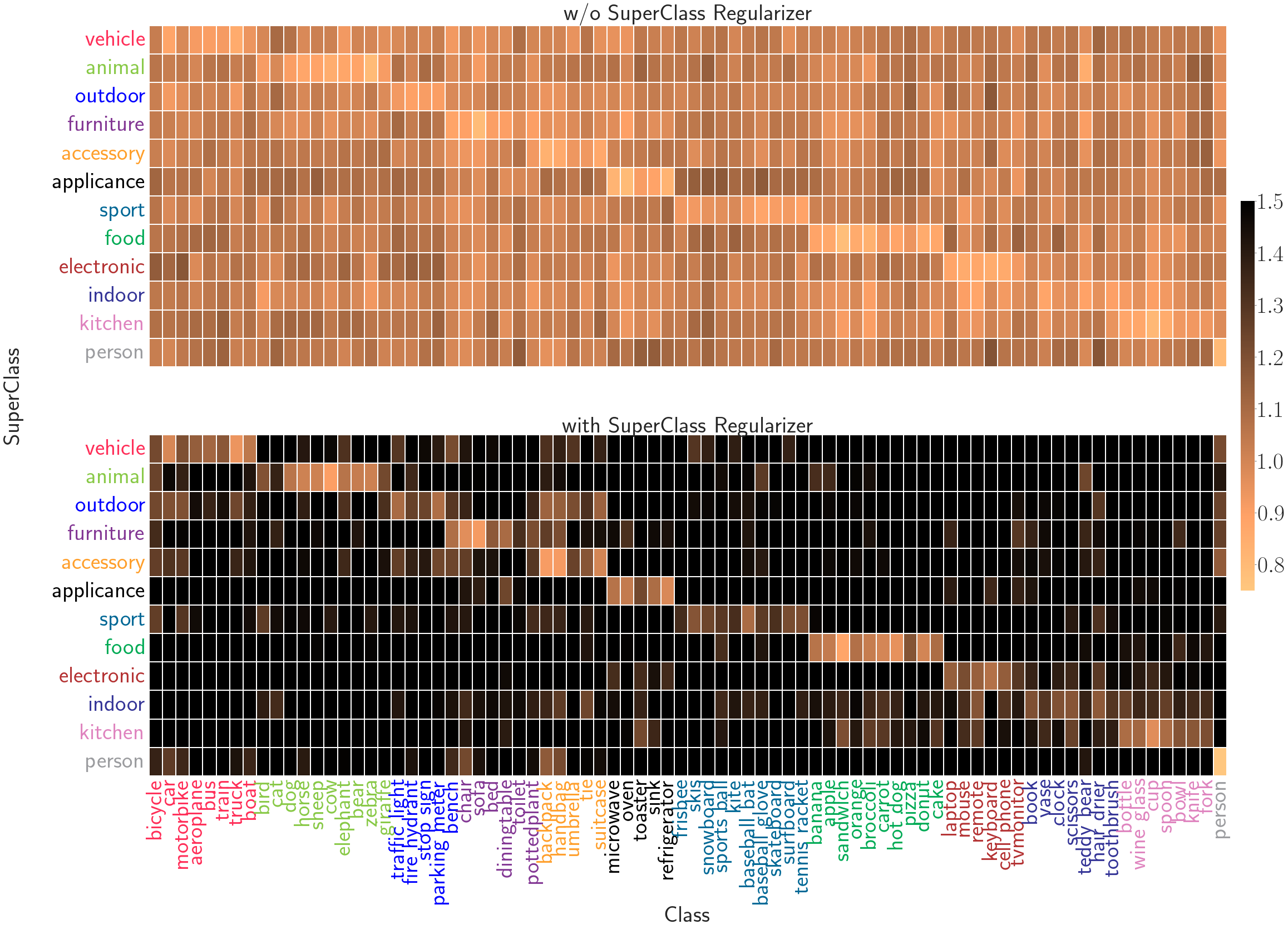}
  \caption{\textbf{Hyperbolic SuperClass-Class Heatmap Distance}.}
  \label{fig:ablation_superclass_regularizer_full}
\end{figure*}

\begin{figure*}[h!]
  % \hspace{-10mm}
  \begin{center}
    \includegraphics[width=0.6\linewidth]{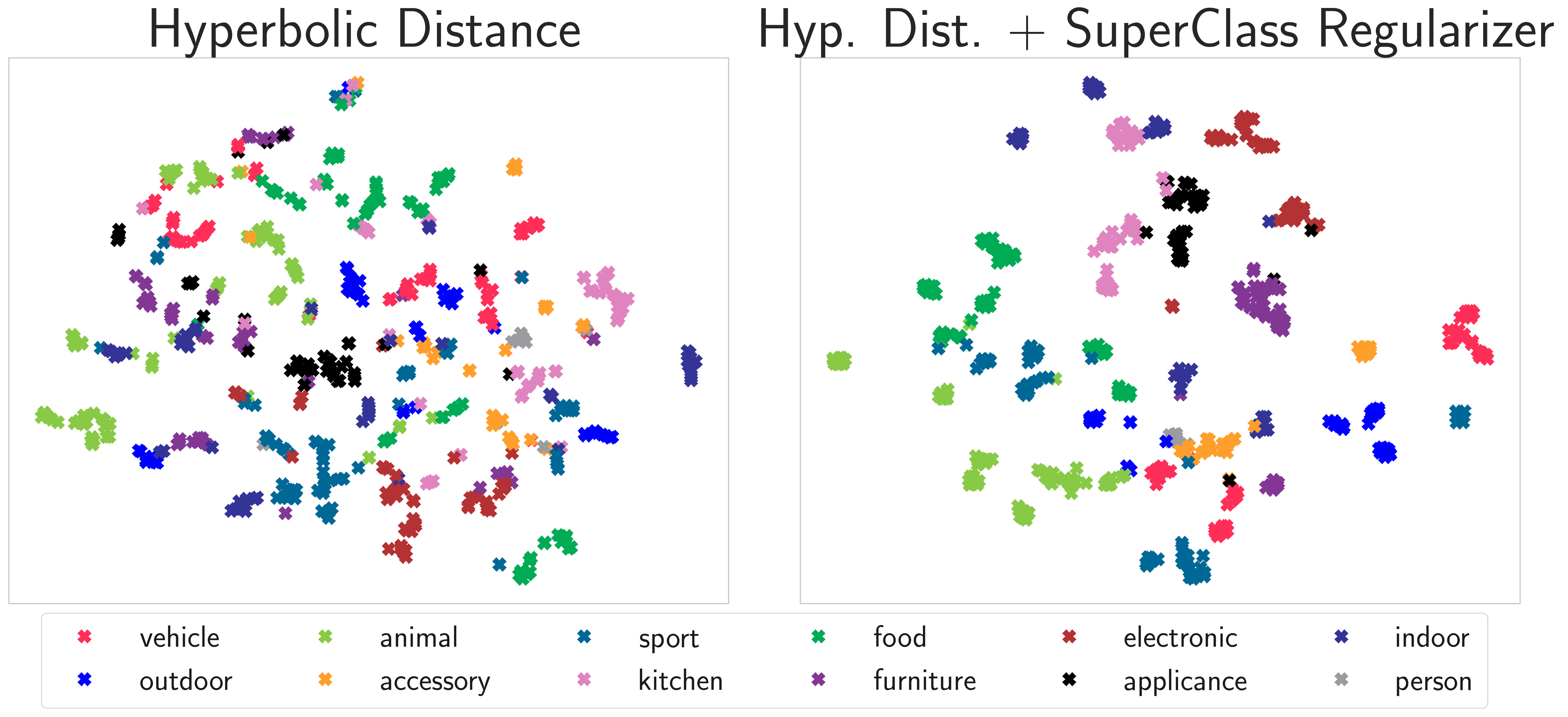}
  \end{center}
  % \vspace{-12mm}
  \caption{\textbf{t-SNE plot of the learned class representations, with colors representing their respective categories.}}
  \label{fig:tse_mini2}
\end{figure*}

Figure~\ref{fig:ablation_superclass_regularizer_full} illustrates the Hyperbolic Distance between each SuperClass (left) and class (bottom) with (top) and without (bottom) our regularizer. The inclusion of the regularizer results in a wider range of values (from $0.7$ to $2.30$) compared to the absence of the regularizer, where the range is narrower (from $0.78$ to $1.2$). This demonstrates that our regularizer effectively pushes classes from different categories apart while bringing classes from the same category closer together. Additionally, without the regularizer, the values are compressed, resulting in a squeezed inter-category distance, as evidenced by the lighter colors and weaker color contrast in the top plot.  T-sne visualization ( Figure~\ref{fig:tse_mini2}) shows a qualitative visualization of this phenomenon.

\clearpage
\subsection{Cosine versus Hyperbolic Distance}
\begin{figure*}[h!]
  \begin{center}
    \includegraphics[width=0.8\textwidth]{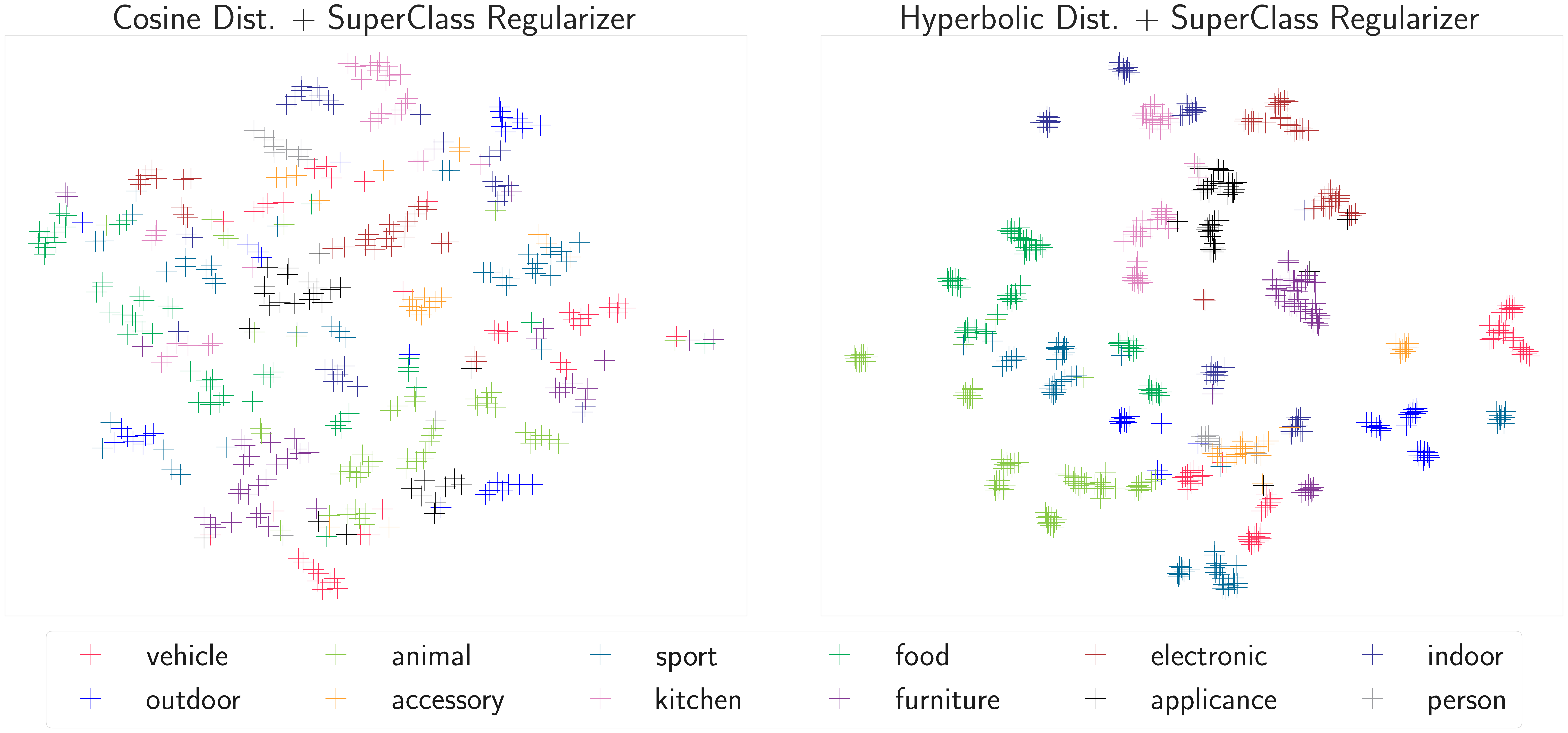}
    % \rule{3cm}{7cm}
  \end{center}
   % \vspace{+32mm}
  \caption{\textbf{t-SNE plot of the learned class embedding where each color represents the class' category.}}
  \label{fig:tsne_cosine_vs_hyperbolic_with_regularizer}
\end{figure*}

Figure~\ref{fig:tsne_cosine_vs_hyperbolic_with_regularizer} displays a t-SNE plot of the learned representation, where each class is color-coded based on its category. While both representations may initially appear scattered, the hyperbolic (right) distance-based representation exhibits improved grouping of classes from the same category. On the other hand, the cosine distance-based representation (left) lacks clear clustering properties between categories, as seen in the appliance category (black), animal category (light green), and furniture category (purple).

These observations highlight the advantage of utilizing hyperbolic distance for capturing meaningful category hierarchy relationship in the embedding space.

\subsection{Impact of Curvature Coefficient c}
To investigate the influence of the curvature coefficient on the hyperbolicity of the embedding space, we conducted experiments using different values of $c$. Our focus was on Task 1 of the Hierarchical Split, where we assessed the model's performance for four values of $c$: $0.0$, $0.1$, $0.2$, and $0.5$. The results are presented in Table~\ref{tab:ablation_curvature}. Notably, the curvature coefficient primarily impacts the U-Recall metric, while the mAP metric remains relatively stable. This observation is expected since the curvature coefficient $c$ directly influences the learned hierarchical structure (hence the  Semantic Similarity Distance), which is the cornerstone of our Adaptive Relabeling Scheme for unknown retrieval. Among the tested values, we found that  $c=0.1$ yielded the optimal performance, consistent with previous findings in the literature~\citep{ermolov2022hyperbolic,hyperbolic_average,yue2023hyperbolic}.

\begin{table*}[ht!]
\centering
\resizebox{0.6\textwidth}{!}{
\begin{tabular}
    {l|cc}  \hline
       &  U-Recall ($\uparrow$)  & mAP($\uparrow$)   \\ \hline
$c=0$ (Cosine Distance)        &  32.8  & 49.0     \\   \hline
   $c=0.1$ (\method)                 &  \textbf{34.9}  & \textbf{49.9}     \\ \hline
   $c=0.2$               &  33.3  &  49.5    \\ \hline
  $c=0.5$                 &  32.3 & 49.8     \\\hline  
\end{tabular}
}
\caption{\textbf{Impact of curvature coefficient $c$ for Hierarchical Split Task 1.}}
\end{table*}